\documentclass{article}

\usepackage{microtype}
\usepackage{graphicx}
\usepackage{booktabs} % for professional tables

\usepackage{hyperref}

\usepackage{url}

\usepackage{subcaption}
\usepackage{placeins}
\usepackage{amssymb}
\usepackage{amsmath}
\usepackage{placeins}
\usepackage{mathtools}
\usepackage{amsthm}
\usepackage{wrapfig}
\usepackage{centernot}
\usepackage{tikz}
\usetikzlibrary{shapes,decorations,arrows,calc,arrows.meta,fit,positioning}
\tikzset{
    -Latex,auto,node distance =1 cm and 1 cm,semithick,
    state/.style ={ellipse, draw, minimum width = 0.7 cm},
    point/.style = {circle, draw, inner sep=0.04cm,fill,node contents={}},
    bidirected/.style={Latex-Latex,dashed},
    el/.style = {inner sep=2pt, align=left, sloped}
}
\usepackage[]{algorithm2e}

\usepackage{booktabs}
\usepackage{siunitx}

\newcommand{\R}{\mathbb{R}}
\newcommand{\E}{\mathbb{E}}
\newcommand{\X}{\mathbf{X}}
\newcommand{\A}{\mathbf{A}}
\newcommand{\B}{\mathbf{B}}
\newcommand{\C}{\mathbf{C}}
\newcommand{\M}{\mathbf{M}}
\newcommand{\HB}{h(\X)}
\newcommand{\h}{h(\x)}
\newcommand{\WX}{\widetilde{\mathbf{X}}}
\newcommand{\x}{\mathbf{x}}

\newcommand{\IL}{\mathcal{L}}

\newcommand{\IE}{\mathcal{E}}
\newcommand{\IES}{\mathcal{E}_{\text{seen}}}
\newcommand{\IEU}{\mathcal{E}_\text{unseen}}

\DeclareMathOperator*{\argmin}{argmin} 
\DeclareMathOperator*{\argmax}{argmax} 

\theoremstyle{plain}
\newtheorem{theorem}{Theorem}
\theoremstyle{plain}
\newtheorem{lemma}{Lemma}
\newtheorem{prop}{Proposition}
\theoremstyle{definition}
\newtheorem{assumption}{Assumption}
\newtheorem{definition}{Definition}
\theoremstyle{remark}

\makeatletter
\theoremstyle{plain}
\newtheorem{repeattheorem@}{Theorem}

\newtheorem{repeatprop@}{Proposition}
\newenvironment{repeatprop}[1]{%
    \def\therepeatprop@{\ref{#1}}
    \repeatprop@
}
{\endrepeatprop@}

\newtheorem{repeatlemma@}{Lemma}
\newenvironment{repeatlemma}[1]{%
    \def\therepeatlemma@{\ref{#1}}
    \repeatlemma@
}
{\endrepeatlemma@}
\makeatother

\usepackage[accepted]{icml2021}

\icmltitlerunning{Learning Robust Models }%Using the Principle of Independent Causal Mechanisms}

\begin{document}

\twocolumn[
\icmltitle{Learning Robust Models Using the \\
Principle of Independent Causal Mechanisms}




\begin{icmlauthorlist}
\icmlauthor{Jens Müller}{hd}
\icmlauthor{Robert Schmier}{hd,bcai}
\icmlauthor{Lynton Ardizzone}{hd}
\icmlauthor{Carsten Rother}{hd}
\icmlauthor{Ullrich Köthe}{hd}
\end{icmlauthorlist}

\icmlaffiliation{hd}{Heidelberg University, Germany}
\icmlaffiliation{bcai}{Bosch Center for Artificial Intelligence, Renningen, Germany}

\icmlcorrespondingauthor{Jens Müller}{jens.mueller@iwr.uni-heidelberg.de\vspace{-6.5mm}}

\icmlkeywords{Machine Learning, ICML}

\vskip 0.3in
]



\printAffiliationsAndNotice{}  % leave blank if no need to mention equal contribution

\begin{abstract}
  Standard supervised learning breaks down under data distribution shift. However, the principle of independent causal mechanisms (ICM, \citet{peters2017elements}) can turn this weakness into an opportunity: one can take advantage of distribution shift between different environments during training in order to obtain more robust models. We propose a new gradient-based learning framework whose objective function is derived from the ICM principle. We show theoretically and experimentally that neural networks trained in this framework focus on relations remaining invariant across environments and ignore unstable ones. Moreover, we prove that the recovered stable relations correspond to the true causal mechanisms under certain conditions. In both regression and classification, the resulting models generalize well to unseen scenarios where traditionally trained models fail.
\end{abstract}

\section{Introduction}
%%%%% Motivation

Standard supervised learning has shown impressive results when training and test samples follow the same distribution. 
However, many real world applications do not conform to this setting, so that research successes do not readily translate into practice \citep{lake2017building}. 
The task of \emph{Domain Generalization} (DG) addresses this problem:
it aims at training models that generalize well under domain shift. 
In contrast to domain \emph{adaption}, where a few labeled and/or many unlabeled examples are provided for each target test domain, in DG
absolutely no data is available from the test domains' distributions 
making the problem unsolvable in general.

\setcounter{footnote}{2} 
In this work, we view the problem of DG specifically using ideas from causal discovery. 
This viewpoint makes the problem of DG well-posed: 
we assume that there exists a feature vector $h^\star(\X)$\footnote{Features can be selected variables or extracted features} 
whose relation to the target variable $Y$ is invariant across all environments.
Consequently, the conditional probability $p(Y\mid h^\star(\X))$ has predictive power in each environment. 
From a causal perspective, changes between domains or environments can be described as interventions; 
and causal relationships -- unlike purely statistical ones -- remain invariant across environments unless explicitly changed under intervention.
This is due to the fundamental principle of ``Independent Causal Mechanisms" which will be discussed in Section \ref{sec:preliminaries}. 
From a causal standpoint, finding robust models is therefore a \emph{causal discovery} task \citep{bareinboim2016causal, meinshausen2018causality}.
Taking a causal perspective on DG, we aim at identifying features which (i) have an invariant relationship to the target variable $Y$ and (ii) are maximally informative about $Y$. 

This problem has already been addressed with some simplifying assumptions and a discrete combinatorial search by \citet{magliacane2018domain, rojas2018invariant}, but we make weaker assumptions and use gradient based optimization. 
The later is attractive because it readily scales to high dimensions and offers the possibility to {\em learn} very informative features, instead of merely selecting among predefined ones. 
Approaches to invariant relations similar to ours were taken by \citet{ghassami2017learning}, who restrict themselves to linear relations, and \citet{arjovsky2019invariant,krueger2020out}, who minimize an invariant empirical risk objective.

Problems (i) and (ii) are quite intricate because the search space has combinatorial complexity and 
testing for conditional independence in high dimensions is notoriously difficult. Our main contributions to this problem are the following:

\begin{itemize}
\item 
By connecting invariant (causal) relations with normalizing flows, we propose a differentiable two-part objective of the form $I(Y; h(\X)) + \lambda_I \mathcal{L}_{I}$, where $I$ is the mutual information and $\mathcal{L}_{I}$ enforces the invariance of the relation between $h(\X)$ and $Y$ across all environments. 
This objective operationalizes the ICM principle with a trade-off between feature informativeness and invariance controlled by parameter $\lambda_I$. 
Our formulation generalizes existing work because our objective is not restricted to linear models. 
\item We take advantage of the continuous objective in three important ways: (1) We can learn invariant new features, whereas graph-based methods as in e.g.~\citet{magliacane2018domain} can only select features from a pre-defined set. (2) Our approach does not suffer from the scalability problems of combinatorial optimization methods as proposed in e.g. \citet{peters2016causal} and \citet{rojas2018invariant}. (3) Our optimization via normalizing flows, i.e. in the form of a density estimation task, facilitates accurate maximization of the mutual information.
\item We show how our objective simplifies in important special cases and under which conditions its optimal solution identifies the true causal parents of the target variable $Y$. We empirically demonstrate that the new method achieves good results on two datasets proposed in the literature. 

\end{itemize}

\vspace{-2mm}
\section{Related Work}
\vspace{-2mm}

Different types of invariances have been considered in the field of DG. 
One type is defined on the feature level, i.e. features $h(\X)$ are invariant across environments if they follow the same distribution in all environments (e.g. \cite{ben2007analysis, pan2010domain, ganin2016domain}).
However, this form of invariance is problematic since for instance the distribution of the target variable might change between environments.
In this case we might expect that the distribution $h(\X)$ changes as well.
A more plausible and theoretically justified type of invariance is the invariance of relations \citep{peters2016causal, magliacane2018domain, rojas2018invariant}.
A relation between a target $Y$ and some features is invariant across environments, if the conditional distribution of $Y$ given the features is the same for all environments.
Existing approaches model a conditional distribution for each feature selection and check for the invariance property  \citep{peters2016causal, rojas2018invariant, magliacane2018domain}. However, this does not scale well. We provide a theoretical result connecting \emph{normalizing flows} and \emph{invariant relations} which in turn allows for gradient-based learning of the problem.
In order to exploit our formulation, we also use the Hilbert-Schmidt-Independence Criterion that has been used for robust learning by \citet{greenfeld2019robust} in the one environment setting.
\citet{arjovsky2019invariant} propose a gradient-based learning framework which exploits a weaker notion of invariance.
Their definition is only a necessary condition, but does not guarantee the more causal definition of invariance we treat in this work.
The connection between DG, invariances and causality has been pointed out for instance by \citet{ zhang2015multi,meinshausen2018causality, rojas2018invariant}.
From a causal perspective, DG is a causal discovery task \citep{meinshausen2018causality}.

For studies on causal discovery in the purely observational setting see e.g. \citet{spirtes1991algorithm,chickering2002optimal, pearl2009causality}, but they cannot take advantage of variations across environments.
The case of different environments has been studied by \citet{hoover1990logic, tian2001causal,mooij2016joint,  peters2016causal, bareinboim2016causal, magliacane2018domain}; \cite{ghassami2018multi, huang2020causal}. Most of these approaches rely on combinatorial methods based on graphical models or are restricted to linear mechanisms, whereas our model defines a continuous objective for very general non-linear models.
The distinctive property of causal relations to remain invariant across environments in the absence of direct interventions has been known since at least the 1930s \citep{frisch1938statistical, heckman2013causal}.
However, its crucial role as a tool for causal discovery was --  to the best of our knowledge-- only recently recognized by \citet{peters2016causal}.
Their estimator -- \emph{Invariant Causal Prediction} (ICP) -- returns the intersection of all subsets of variables that have an invariant relation w.r.t. $Y$. 
The output is shown to be the set of the direct causes of $Y$ under suitable conditions.
However, their method assumes an underlying linear model and must perform an exhaustive search over all possible variable sets $\X_S$, which does not scale.
Extensions to time series and non-linear additive noise models were studied in  \citet{heinze2018invariant, pfister2019invariant}.
Our treatment of invariance is inspired by these papers and also discusses identifiability results, i.e. conditions when the identified variables are indeed the direct causes.
Key differences between ICP and our approach are the following:
Firstly, we propose a formulation that allows for a gradient-based learning without strong assumptions on the underlying causal model such as linearity. 
Second, while ICP tends to exclude features from the parent set when in doubt, our algorithm prefers to err in the direction of best prediction performance in this case.

\vspace{-2mm}
\section{Preliminaries}
\vspace{-2mm}
%%%%% Preliminaries 
\label{sec:preliminaries}

In the following we introduce the basics of this article as well as the connection between DG and causality. Basics on causality are presented in Appendix \ref{app:causalbasics}. 
We first define our notation as follows: We denote the set of all variables describing the system under study as $\widetilde{\X} = \{X_1, \dots, X_D\}$.
One of these variables will be singled out as our prediction target, whereas the remaining ones are observed and may serve as predictors.
To clarify notation, we call the target variable $Y \equiv X_i$ for some $i \in \{1, \dots, D \}$, and the remaining observations are $\X = \widetilde{\X} \setminus \{Y\}$. Realizations of a random variable are denoted with lower case letters, e.g. $x_i$. 
We assume that observations can be obtained in different environments $e \in \IE$.
Symbols with superscript, e.g. $Y^e$, refer to a specific environment, whereas symbols without refer to data pooled over all environments.
We distinguish known environments $e \in \IES$, where training data are available, from unknown ones $e\in \IEU$, where we wish our models to generalize to.
The set of all environments is $\IE = \IES \cup \IEU$.
We assume that all random variables have a density $p_A$ with probability distribution $P_A$ (for some variable or set $A$). We consider the environment to be a random variable $E$ and therefore a system variable similar to \citet{mooij2016joint}. This gives an additional view on casual discovery and the DG problem.

Independence and dependence of two variables $A$ and $B$ is written as $A \perp B$ and $A\not \perp B$ respectively. 
Two random variables $A,B$ are conditionally independent given $C$ if  $P(A, B\mid C) = P(A\mid C) P(B \mid C)$. 
This is denoted with $A \perp B \mid C$. 
Intuitively, it means $A$ does not contain any information about $B$ if $C$ is known \citep[for details see e.g. ][]{peters2017elements}. Similarly, one can define independence and conditional independence for sets of random variables.

\subsection{Invariance and the Principle of ICM}

DG is in general unsolvable because distributions between seen and unseen environments could differ arbitrarily. 
In order to transfer knowledge from $\IES$ to $\IEU$, we have to make assumptions on how seen and unseen environments relate. These assumptions have a close link to causality.

We assume certain relations between variables remain invariant across all environments.  
A subset $\X_S \subset \X$ of variables \emph{elicits an invariant relation} or \emph{satisfies the invariance property} w.r.t. $Y$ over a subset 
$W \subset \IE$ of environments if 
\begin{align}
        \label{eq:invarianceCondSets}
         \forall e, e' \in W\colon\quad P(Y^e \mid \X_S^e = u) = P(Y^{e'} \mid \X_S^{e'}=u)
\end{align}
for all $u$ where both conditional distributions are well-defined. 
Equivalently, we can define the invariance property by $Y \perp E \mid \X_S$ and $I(Y;E \mid \X_S) = 0$ for $E$ restricted to $W$. The  \emph{invariance property} for computed features $h(\X)$ is defined analogously by the relation $Y \perp E \mid h(\X)$. 

Although we can only test for \eqref{eq:invarianceCondSets} in $\IES$, taking a causal perspective allows us to derive plausible conditions 
for an invariance to remain valid in all environments $\IE$. 
In brief, we assume that environments correspond to interventions in the system and invariance arises from the principle of {\em independent causal mechanisms} \citep[ICM]{peters2017elements}. We specify these conditions later in Assumption \ref{assum:MainAssumption1} and \ref{assum:MainAssumption2}.

At first, consider the joint density $p_{\WX}(\WX)$.
The chain rule offers a combinatorial number of ways to decompose this distribution into a product of conditionals.
Among those, the {\em causal factorization}
\begin{align}
    \label{eq:causalFactorization}
    p_{\WX} (x_1, \dots, x_D) = \prod_{i=1}^D p_i ( x_i \mid \x_{pa(i)})
\end{align}
is singled out by conditioning each $X_i$ onto its {\em direct causes} or {\em causal parents} $\X_{pa(i)}$, where $pa(i)$ denotes the appropriate index set.
The special properties of this factorization are discussed in \citet{peters2017elements}. 
The conditionals $p_i$ of the causal factorization are called {\em causal mechanisms}.
An {\em intervention} onto the system is defined by replacing one or several factors in the decomposition with different (conditional) densities $\overline{p}$.
Here, we distinguish \emph{soft-interventions} where $\overline{p}_j(x_j \mid \x_{pa(j)}) \ne p_j(x_j \mid \x_{pa(j)})$ and \emph{hard}-interventions where $\overline{p}_j(x_j \mid \x_{pa(j)}) = \overline{p}_j(x_j)$ is a density which does not depend on $x_{pa(j)}$ (e.g. an atomic intervention where $x_j$ is forced to take a specific value $\overline{x}$). 
The resulting joint distribution for a single intervention is
\begin{align}
    \label{eq:interventionalFactorization}
    \overline{p}_{\WX} (x_1, \dots, x_D) = \overline{p}_j (x_j \mid \x_{pa(j)}) \prod_{i=1, i\ne j}^D p_i ( x_i \mid \x_{pa(i)})
\end{align}
and extends to multiple simultaneous interventions in the obvious way. 
The principle of {\em independent causal mechanisms} (ICM) states that every mechanism acts independently of the others \citep{peters2017elements}. 
Consequently, an intervention replacing $p_j$ with $\overline{p}_j$ has no effect on the other factors $p_{i\ne j}$, as indicated by \eqref{eq:interventionalFactorization}.
This is a crucial property of the causal decomposition -- alternative factorizations do not exhibit this behavior.
Instead, a coordinated modification of several factors is generally required to model the effect of an intervention in a non-causal decomposition.

We utilize this principle as a tool to train {\em robust} models.
To do so, we make two additional assumptions, similar to \citet{peters2016causal} and \citet{heinze2018invariant}:

\begin{assumption}
\label{assum:MainAssumption1}
Any differences in the joint distributions $p^e_{\WX}$ from one environment to the other are fully explainable as interventions: replacing factors $p_i^e ( x_i \mid \x_{pa(i)})$ in environment $e$ with factors $p_i^{e'} ( x_i \mid \x_{pa(i)})$ in environment $e'$ (for some subset of the variables) is the only admissible change.
\end{assumption}
\begin{assumption}
\label{assum:MainAssumption2}
The mechanism $p( y \mid \x_{pa(Y)})$ for the target variable is invariant under changes of environment. In other words, we require conditional independence $Y \perp E \mid \X_{pa(Y)}$.
\end{assumption}

Assumption 2 implies that $Y$ must not directly depend on $E$. 
In addition, it has important consequences when there exist omitted variables $\mathbf{W}$, which influence $Y$ but have not been measured.
Specifically, if the omitted variables depend on the environment (hence $\mathbf{W} \centernot{\perp} E$) or $\mathbf{W}$ contains a hidden confounder of $\X_{pa(Y)}$ and $Y$ while $\X_{pa(Y)} \not \perp E$ (the system is not causally sufficient and $\X_{pa(Y)}$ becomes a ``collider'', hence $\mathbf{W}\centernot{\perp} E \mid \X_{pa(Y)}$), then $Y$ and $E$ are no longer $d$-separated by $\X_{pa(Y)}$ and Assumption 2 is unsatisfiable.
Then our method will be unable to find an invariant mechanism (see Appendix \ref{app:discussAssumptions} for more details).

If we knew the causal decomposition, we could use these assumptions directly to train a robust model for $Y$ -- we would simply regress $Y$ on its parents $\X_{pa(Y)}$.
However, we only require that a causal decomposition with these properties exists, but do not assume that it is known.
Instead, our method uses the assumptions indirectly 

-- by simultaneously considering data from different environments -- to identify a stable regressor for $Y$. 

 We call a regressor stable if it solely relies on predictors whose relationship to $Y$ remains invariant across environments, i.e. is not influenced by any intervention.
By assumption \ref{assum:MainAssumption2}, such a regressor always exists.
However, predictor variables beyond $\X_{pa(Y)}$ may be used as well, e.g. children of $Y$ or parents of children, provided their relationships to $Y$ do not depend on the environment.
The case of children is especially interesting: Suppose $X_j$ is a noisy measurement of $Y$, described by the causal mechanism $P(X_j\mid Y)$.
As long as the measurement device works identically in all environments, including $X_j$ as a predictor of $Y$ is desirable, despite it being a child. We discuss and illustrate Assumption \ref{assum:MainAssumption2} in Appendix \ref{app:discussAssumptions}. In general, prediction accuracy will be maximized when all suitable predictor variables are included into the model.
Accordingly, our algorithm will asymptotically identify the full set of stable predictors for $Y$.
In addition, we will prove under which conditions this set contains exactly the parents of $Y$.
Note that there are different ideas on whether most supervised learning tasks conform to this setting \citep{scholkopf2012causal, arjovsky2019invariant}.

\subsection{Domain Generalization}
In order to exploit the principle of ICM for DG, we formulate the DG problem as follows
\begin{align}
        \label{eq:ourEstimator2} 
         h^{\star} &\coloneqq \argmax_{h\in \mathcal{H}} \Big\{ \min_{e \in \IE} I(Y^e; h(\X^e)) \Big\} \nonumber \\ & \quad\text{s.t.}\quad Y \perp E \mid h(\X)  
\end{align}
where $h\in \mathcal{H}$ denotes a learnable feature extraction function $h \colon \R^D \to \R^M$ where $M$ is a hyperparameter. 
This optimization problem defines a maximin objective: 
The features $h(\X)$ should be as informative as possible about the response $Y$ even in the most difficult environment, 
while conforming to the ICM constraint
that the relationship between features and response must remain invariant across all environments. 
In principle, our approach can also optimize related objectives like the average mutual information over environments.
However, very good performance in a majority of the environments could then mask failure in a single (outlier) environment. 
We opted for the maximin formulation to avoid this.

As it stands, \eqref{eq:ourEstimator2} is hard to optimize, 
because traditional independence tests for the constraint $ Y \perp E \mid h(\X)$ cannot cope with conditioning variables selected from a potentially infinitely large space $\mathcal{H}$. 
A re-formulation of the DG problem to circumvent these issues is our main theoretical contribution. 
\newline

\subsection{Normalizing Flows}
\label{app:normflow}

Normalizing flows form a class of probabilistic models that has recently received considerable attention, see e.g. \citet{papamakarios2019normalizing} for an in-depth review for 
Appendix \ref{app:normFlows}.
They model complex distributions by means of invertible functions $T$ (chosen from some model space $\mathcal{T}$) which map the densities of interest to latent normal distributions. 
The inverses $F=T^{-1}$ then act as generative models for the target distributions.
Normalizing flows are typically built with specialized neural networks that are invertible by construction and have tractable Jacobian determinants.

In our case, we represent the conditional distribution $P(Y\mid h(\X))$ using a {\em conditional} normalizing flow \citep[see e.g.][]{ardizzone2019guided}.
To this end, we seek a mapping $R=T(Y; h(\X))$ that is diffeomorphic in $Y$ such that $R\sim \mathcal{N}(0,1) \perp h(\X)$ when $Y \sim P(Y\mid h(\X))$. 
This is a generalization of the well-studied additive Gaussian noise model $R=Y-f(h(\X))$, see Section \ref{sec:seploiting-invariances}.
The inverse $Y=F(R; h(\X))$ assumes the role of a structural equation for the mechanism $p(Y\mid h(\X))$ with $R$ being the corresponding noise variable.
\footnote{$F$ is the concatenation of the normal CDF with the inverse CDF of $P(Y\mid h(\X))$, see \citet{peters2014causal}.}
However, in our context it is most natural to learn $T$ (rather than $F$) by minimizing the negative log-likelihood (NLL) of $Y$ under $T$ \citep{papamakarios2019normalizing},
which takes the form
\begin{align}
\label{eq:nll-objective}
\mathcal{L}_{\mathrm{NLL}}  (T, h) \coloneqq & \ \E_{h(\X),Y} \big[\|T(Y; h(\X) \|^2/2 \notag\\ & - \log |\det \nabla_y T(Y;h(\X))|\big]+C
\end{align}
where $\det \nabla_y T$ is the Jacobian determinant and $C= \dim(Y) \log(\sqrt{2\pi})$ is a constant that can be dropped. 
If we consider the NLL on a particular environment $e \in \IE$, we denote this with $\mathcal{L}^e_\mathrm{NLL}$.
Lemma \ref{lem:normalizingFlows} shows that normalizing flows optimized by NLL are indeed applicable to our problem:
\begin{lemma}
\label{lem:normalizingFlows}
(proof in Appendix \ref{app:normFlows}) Let $  h^{\star}, T^{\star} \coloneqq \arg \min_{h \in \mathcal{H}, T \in \mathcal{T}} \mathcal{L}_{\mathrm{NLL}}  (T, h)$ be the solution of the NLL minimization problem on a sufficiently rich function space $\mathcal{T}$. Then the following properties are guaranteed for arbitrary sets $\mathcal{H}$ of feature extractors:
\begin{itemize}
    \item[(a)] $h^{\star}$ also maximizes the mutual information, i.e. $h^{\star} = g^{\star}$ with $g^{\star}= \arg \max_{g \in \mathcal{H}} I(g(\X); Y)$
    \item[(b)] $h^{\star}$ is independent of the flow's latent variable: $h^{\star}(\X) \perp  R$ with $R=T^{\star}(Y; h^{\star}(\X))$.
\end{itemize}
\end{lemma}
Statement (a) guarantees that $h^\star$ extracts as much information about $Y$ as possible.
Hence, the objective (\ref{eq:ourEstimator2}) becomes equivalent to optimizing (\ref{eq:nll-objective}) when we restrict the space $\mathcal{H}$ of admissible feature extractors to the subspace $\mathcal{H}_{\perp }$ satisfying the invariance constraint $Y \perp E \mid h(\X)$:  $\argmin_{h \in \mathcal{H}_\perp}\max_{e \in \IE}\min_{ T\in \mathcal{T}}  \mathcal{L}_\mathrm{NLL}^e(T;h) = \argmax_{h \in \mathcal{H}_\perp} \min_{e \in \IE} I(Y^e; h(\X^e))$ (Appendix \ref{app:normFlows}).
Statement (b) ensures that the flow indeed implements a valid structural equation, which requires that $R$ can be sampled independently of the features $h(\X)$.

\section{Method}
\label{sec:methods}

In the following we propose a way of indirectly expressing the constraint in \eqref{eq:ourEstimator2} via normalizing flows. 
Thereafter, we combine this result with Lemma \ref{lem:normalizingFlows} to obtain a differentiable objective for solving the DG problem. 
We also present important simplifications for least squares regression and softmax classification and discuss relations of our approach with causal discovery.

\subsection{Learning the Invariance Property}

The following theorem establishes a connection between invariant relations, prediction residuals and normalizing flows.
The key consequence is that a suitably trained normalizing flow translates the statistical independence of the latent variable $R$ from the features and environment $(h(\X), E)$ into the desired invariance of the mechanism $P(Y \mid h(\X))$ under changes of $E$. 
We will exploit this for an elegant reformulation of the DG problem (\ref{eq:ourEstimator2}) into the objective (\ref{eq:flow-and-hsic-loss}) below.

\begin{theorem}
\label{thm:residualTheorem}
Let $h$ be a differentiable function and $Y, \X, E$ be random variables. Furthermore, let $R=T(Y; h(\X))$ be a continuous, differentiable function that is a diffeomorphism in $Y$. Suppose that $R \perp (h(\X), E)$. 
Then, it holds that $Y \perp E \mid h(\X)$.
\end{theorem}
\begin{proof}
The decomposition rule for the assumption $R \perp (h(\X), E)$ (i) implies $R \perp h (\X)$ (ii).
To simplify notation, we define $Z:=h(\X)$. Because $T$ is invertible in $Y$ and due to the change of variables (c.o.v.) formula, we obtain 
\begin{align*}
    p_{Y\mid Z,E}(y \mid z,e)
    \overset{(c.o.v.)}{=} &p_{R \mid Z, E}( T(y, z) \mid z, e ) \left |\det \frac{\partial T}{\partial y} (y, z) \right | \\ \overset{(i)}{=} \;\; &p_R(r)  
    \left|\det \frac{\partial T}{\partial y} (y, z) \right | \\
    \overset{(ii)}{=}\;\; &p_{R \mid Z} (r  \mid z)\left|\det \frac{\partial T}{\partial y}  (y,z )\right | \\
    \overset{(c.o.v.)}{=} &p_{Y \mid Z}(y\mid z).
\end{align*}
This implies $Y \perp E \mid Z$.
\end{proof}

The theorem states in particular that if there exists a suitable diffeomorphism $T$ such that $R \perp (h(\X), E)$, then $h(\X)$ satisfies the invariance property w.r.t. $Y$. 
Note that if Assumption 2 is violated, the condition $R \perp (h(\X), E)$ is unachievable in general and therefore the theorem is not applicable (see Appendix \ref{app:discussAssumptions}).
We use Theorem \ref{thm:residualTheorem} in order to {\em learn} features $h$ that meet this requirement. 
In the following, we denote a conditional normalizing flow parameterized via $\theta$ with $T_\theta$. Furthermore, $h_\phi$ denotes a feature extractor implemented as a neural network parameterized via $\phi$.  
We can relax condition $R\perp (h_\phi(\X), E)$ by means of the Hilbert Schmidt Independence Criterion (HSIC), a kernel-based independence measure (see Appendix \ref{app:HSIC} for the mathematical definition and \citet{gretton2005measuring} for details). This loss, denoted as $\IL_I$, penalizes dependence between the distributions of $R$ and $(h_\phi(\X),E)$. The HSIC guarantees that 
\begin{align}
    \label{eq:HSICEquivalence}
       \IL_I\big(P_{R}, P_{h_\phi (\X), E}\big) = 0\quad \Longleftrightarrow\quad  R \perp  (h_\phi(\X) , E)  
\end{align}
where $R = T_\theta(Y ; h_\phi(\X))$ and $P_{R}, P_{h_\phi (\X), E}$ are the distributions implied by the parameter choices $\phi$ and $\theta$. Due to Theorem \ref{thm:residualTheorem}, minimization of $\IL_I (P_R, P_{h_\phi(\X), E})$ w.r.t. $\phi$ and $\theta$ will thus approximate the desired invariance property $Y \perp E \mid h_\phi(\X)$, with exact validity upon perfect convergence.

When $R \perp  (h_\phi(\X) , E)$ is fulfilled, the decomposition rule implies $R \perp  E$ as well. 
However, if the differences between environments are small, empirical convergence is accelerated by adding a Wasserstein loss which explicitly enforces the latter, see Appendix \ref{app:HSIC} and Section \ref{sec:colored-mnist} for details.

\subsection{Exploiting Invariances for Prediction}
\label{sec:seploiting-invariances}

Equation \eqref{eq:ourEstimator2} can be re-formulated as a differentiable loss using a Lagrange multiplier $\lambda_I$ on the HSIC loss. 
$\lambda_I$ acts as a hyperparameter to adjust the trade-off between the invariance property of $h_\phi(\X)$ w.r.t. $Y$ and the mutual information between $h_\phi(\X)$ and $Y$. See Appendix \ref{app:algo} for algorithm details.

\paragraph{Normalizing Flows}

Using Lemma \ref{lem:normalizingFlows}(a), we maximize $\min_{e \in \IE} I(Y^e; h_\phi(\X^e))$ by minimizing 
$ \max_{e \in \IE} \{ \mathcal{L}_{\mathrm{NLL}}(T_\theta; h_\phi) \}$ w.r.t. $\phi, \theta$.  
To achieve the described trade-off between goodness-of-fit and invariance, we therefore optimize
\begin{align}
\label{eq:flow-and-hsic-loss}
      &\arg \min_{\theta, \phi}   \Big( \max_{e \in \IE} \Big \{ \mathcal{L}_{\mathrm{NLL}}(T_\theta, h_\phi)  \Big\} + \lambda_I \IL_I (P_R, P_{h_\phi(\X), E}) \Big)
\end{align}
where 
$R^e = T_\theta (Y^e, h_\phi(\X^e))$ and $\lambda_I >0$. The first term maximizes the mutual information between $h_\phi(\X)$ and $Y$ in the environment where the features are least informative about $Y$ and the second term aims to ensure an invariant relation.

\paragraph{L2-Regression under Additive Noise}

Let $f_\theta$ be a regression function. Solving for the noise term gives $R = Y - f_\theta(\X)$ which corresponds to a diffeomorphism in $Y$, namely $T_\theta(Y; X) = Y- f_\theta(\X)$. If we make two simplified assumptions: (i) the noise is gaussian with zero mean and (ii) $R\perp f_\theta(\X)$, 
then we obtain 
\begin{align*}
I(Y; f_\theta(\X))&= H(Y) - H(Y \mid f_\theta(\X)) \\ 
&= H(Y) - H(R \mid f_\theta(\X)) \\
&\overset{(ii)}{=} H(Y) - H(R) \\
&\overset{(i)}{=}  H(Y) - 1/2 \log (2 \pi e \sigma^2)  
\end{align*} 
where $\sigma^2 = \E[(Y-f_\theta(\X))^2]$. In this case maximizing the mutual information $I(Y;f_\theta(\X))$ amounts to minimizing $\E [(Y- f_\theta(\X))^2]$ w.r.t. $\theta$, i.e. the standard L2-loss for regression problems. 
From this, we obtain a simplified version of \eqref{eq:ourEstimator2} via
\begin{align}
          \arg \min_{\theta} \Big( \max_{e \in \IES} \Big\{ & \E\big[(Y^e-f_\theta(\X^e))^2\big] \Big\} \notag\\  + &\lambda_I \IL_I (P_R, P_{f_\theta(\X), E}) \Big)
              \label{eq:additiveNoiseDG}
\end{align}
where $R^e = Y- f_\theta( \X^e)$ and $\lambda_I > 0$. Under the conditions stated above, the objective achieves the mentioned trade-off between information and invariance.

Alternatively we can view the problem as to find features $h_\phi \colon \R^D \to \R^m$ such that $I(h_\phi(\X), Y)$ gets maximized under the assumption that there exists a model $f_\theta(h_\phi(\X)) + R = Y$ where $R$ is independent of $h_\phi(\X)$ and $R$ is gaussian. In this case we obtain similarly as above the learning objective
\begin{align}
          \arg \min_{\theta, \phi}   \Big( \max_{e \in \IES} \Big\{ & \E\big[(Y^e-f_\theta(h_\phi(\X^e)))^2\big] \Big\} \notag \\  + & \lambda_I \IL_I (P_R, P_{h_\phi(\X), E}) \Big)
    \label{eq:additiveNoiseDG2}
\end{align}

\paragraph{Classification}
The expected cross-entropy loss is given through 
\begin{align}
    - \E_{\X,Y} \Big[f(\X)_Y - \log \Big( \sum_{c} \exp ( f(\X)_c\Big)\Big]
\end{align}
where $f\colon \mathcal{X} \to \R^m$ returns the logits. Minimizing the expected cross-entropy loss amounts to maximizing the mutual information between $f(\X)$ and $Y$ \citep[eq. 3]{qin2019rethinking,barber2003algorithm}. Let $T(Y; f(\X)) = Y\cdot \mathrm{softmax}(f(\X))$ with component-wise multiplication, then $T$ is invertible in $Y$ conditioned on the softmax output. Now we can apply the same invariance loss as above in order to obtain a solution to \eqref{eq:ourEstimator2}.

\subsection{Relation to Causal Discovery}

Under certain conditions, solving \eqref{eq:ourEstimator2} leads to features which correspond to the direct causes of $Y$ (identifiability). In this case, we obtain the causal mechanism by computing the conditional distribution of $Y$ given the direct causes. Therefore \eqref{eq:ourEstimator2} can also be seen as approximation of the causal mechanism when the identifiability conditions are met. The following Proposition states under which assumptions the direct causes of $Y$ can be recovered by exploiting Theorem \ref{thm:residualTheorem}.
\begin{prop}
    \label{prop:identifiability2}
    We assume that the underlying causal graph $G$ is faithful with respect to $P_{\WX,  E}$.
	We further assume that every child of $Y$ in $G$ is also a child of $E$ in $G$. A variable selection $h(\X)= \X_S$ corresponds to the direct causes if the following conditions are met: (i) $T(Y;(X)) \perp E, h(\X)$ is satisfied for a diffeomorphism $T(\cdot ; h(\X))$, (ii) $h(\X)$ is maximally informative about $Y$ and (iii) $h(\X)$ contains only variables from the Markov blanket of $Y$.
\end{prop}
The Markov blanket of $Y$ is the only set of vertices which are necessary to predict $Y$ (see Appendix \ref{app:causalbasics}). We give a proof of Proposition \ref{prop:identifiability2} as well as a discussion in Appendix \ref{app:identifiability}.

For reasons of explainability and for the task of causal discovery, we employ a gating function $h_\phi$ in order to obtain a variable selection. The gating function $h_\phi$ represents a $0$-$1$ mask of the input. A complexity loss $\mathcal{L}(\phi)$ represents how many variables are selected and therefore penalizes to include variables. We use the same gating function and complexity loss as in \citet{kalainathan2018sam}. Intuitively speaking, if we search for a variable selection that conforms to the conditions in Proposition \ref{prop:identifiability2}, the complexity loss would exclude all non-task relevant variables. Therefore, if $\mathcal{H}$ is the set of gating functions, then $h^\star$ in \eqref{eq:ourEstimator2} would correspond to the direct causes of $Y$  under the conditions listed in Proposition \ref{prop:identifiability2}. 
The complexity loss as well as the gating function can be optimized by gradient descent. 

\vspace{-2mm}
\section{Experiments}
\vspace{-2mm}
\label{sec:experiments}

\subsection{Synthetic Causal Graphs}

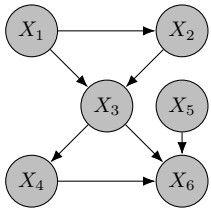
\begin{wrapfigure}{r}{0.25\textwidth}    \centering
\begin{tikzpicture}
	\node[state, circle, scale=0.8, fill=gray!50] (x0) at (-1, 1) {$X_1$};
	\node[state, circle, scale=0.8, fill=gray!50] (x1) at (1,1) {$X_2$};
	\node[state, circle, scale=0.8, fill=gray!50] (x2) at (0, 0) {$X_3$};
	\node[state, circle, scale=0.8, fill=gray!50] (x3) at (-1, -1) {$X_4$};
	\node[state, circle, scale=0.8, fill=gray!50] (x4) at (1, 0) {$X_5$};
	
	\node[state, circle, scale=0.8, fill=gray!50] (x5) at (1, -1) {$X_6$};

	\path (x0) edge (x2);
	\path (x0) edge (x1);
	\path (x1) edge (x2);
	\path (x2) edge (x3);
	\path (x2) edge (x5);
	\path (x4) edge (x5);
	\path (x3) edge (x5);
\end{tikzpicture}
    \caption{Directed graph of our SCM. Target variable $Y$ is chosen among $X_1,\dots, X_6$ in turn.}
    \label{fig:heinze_graph}
\end{wrapfigure}

To evaluate our methods for the regression case, we follow the experimental design of \cite{heinze2018invariant}.
It rests on the causal graph in Figure \ref{fig:heinze_graph}.
Each variable $X_1,...,X_6$ is chosen as the regression target $Y$ in turn, so that a rich variety of local configurations around $Y$ is tested.
The corresponding structural equations are selected among four model types of the form $f(\X_{pa(i)}, N_i) = \sum_{j\in pa(i)} \texttt{mech}( a_j X_j) + N_i$, where $\texttt{mech}$ is either the identity (hence we get a linear Structural Causal Model (SCM)), Tanhshrink, Softplus or ReLU, and one multiplicative noise mechanism of the form $f_{i} ( \X_{pa(i)}, N_i) = (\sum_{j \in pa(i)} a_j X_j ) \cdot ( 1+ (1/4) N_i) +N_i$, resulting in 1365 different settings. 
For each setting, we define an observational environment (using exactly the selected mechanisms) and three interventional ones, where soft or do-interventions are applied to non-target variables according to Assumptions \ref{assum:MainAssumption1} and \ref{assum:MainAssumption2} (full details in Appendix \ref{app:experimentalSetting}).
Each inference model is trained on 1024 realizations of three environments, whereas the fourth one is held back for DG testing. 
The tasks are to identify the parents of the current target variable $Y$, and to train a transferable regression model based on this parent hypothesis.
We measure performance by the accuracy of the detected parent sets and by the L2 regression errors relative to the regression function using the ground-truth parents.

{}

We evaluate four models derived from our theory: two normalizing flows as in \eqref{eq:flow-and-hsic-loss} with and without gating mechanisms (FlowG, Flow) and two additive noise models, again with and without gating mechanism (ANMG, ANM), using a feed-forward network with the objective in \eqref{eq:additiveNoiseDG2} (ANMG) and \eqref{eq:additiveNoiseDG} (ANM).
For comparison, we train three baselines: ICP (a causal discovery algorithm also exploiting ICM, but restricted to linear regression, \citet{peters2016causal}), a variant of the PC-Algorithm (PC-Alg, see Appendix \ref{app:pcVariant}) and standard empirical-risk-minimization ERM, a feed-forward network minimizing the L2-loss, which ignores the causal structure by regressing $Y$ on all other variables. 
We normalize our results with a ground truth model (CERM), which is identical to ERM, but restricted to the true causal parents of the respective $Y$. 

\begin{figure}
\centering

  \includegraphics[trim={12 0 7 0},clip,width=1.0\linewidth]{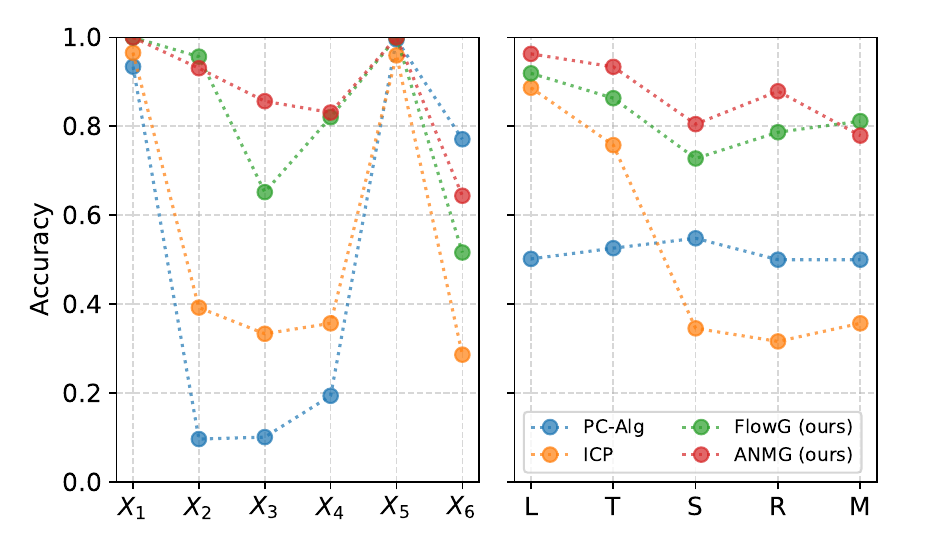}
  \caption{Detection accuracy of the direct causes for baselines and our gating architectures, broken down for different target variables (left) and mechanisms (right: \textbf{L}inear, \textbf{T}anhshrink, \textbf{S}oftplus, \textbf{R}eLU, \textbf{M}ultipl. Noise).}
  \label{fig:acc_mechanisms_exp}
\end{figure}

The accuracy of parent detection is shown in Figure \ref{fig:acc_mechanisms_exp}. The score indicates the fraction of the experiments where the exact set of all causal parents was found and all non-parents were excluded.
We see that the PC algorithm performs unsatisfactorily, whereas 
ICP exhibits the expected behavior: it works well for variables without parents and for linear SCMs, i.e. exactly within its specification.
Among our models, only the gating ones explicitly identify the parents.
They clearly outperform the baselines, with a slight edge for ANMG, as long as its assumption of additive noise is fulfilled.

\begin{figure}
\includegraphics[width=1.\linewidth]{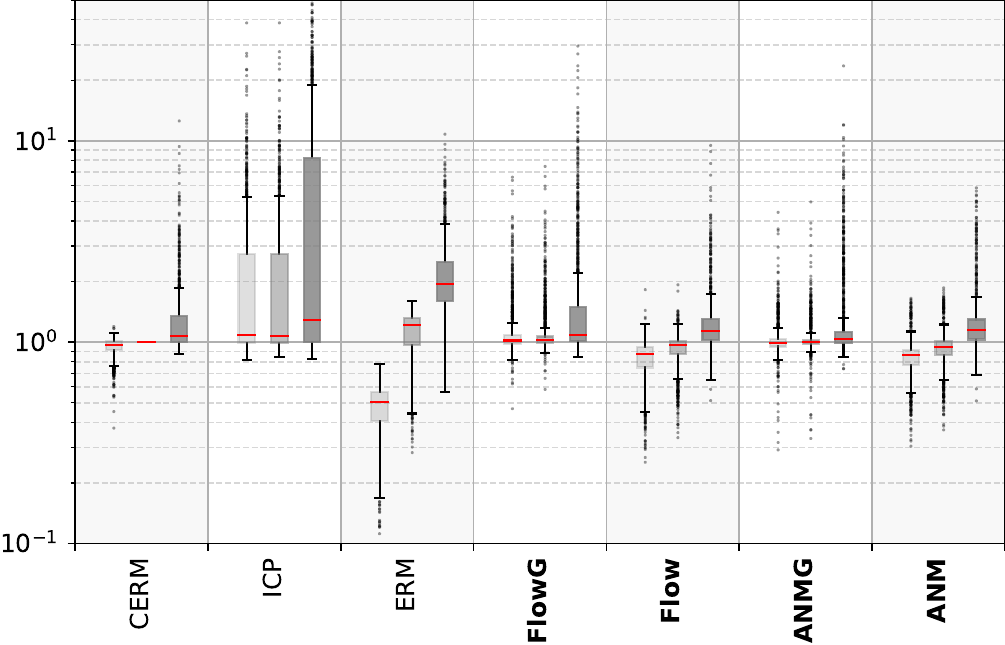}  
  \caption{Logarithmic plot of L2 errors, normalized by CERM test error. For each method (ours in bold) from left to right: training error, test error on seen environments, domain generalization error on unseen environments.}
  \label{fig:transfer_exp}

\end{figure}

Figure \ref{fig:transfer_exp} and Table \ref{tab:oodMechanisms} report regression errors for seen and unseen environments, with CERM indicating the theoretical lower bound. 
The PC algorithm is excluded from this experiment due to its poor detection of the direct causes.
ICP wins for linear SCMs, but otherwise has largest errors, since it cannot accurately account for non-linear mechanisms.
ERM gives reasonable test errors (while overfitting the training data), but generalizes poorly to unseen environments, as expected.
Our models perform quite similarly to CERM.
We again find a slight edge for ANMG, except under multiplicative noise, where ANMG's additive noise assumption is violated and Flow is superior.
All methods (including CERM) occasionally fail in the domain generalization task, indicating that some DG problems are more difficult than others, e.g. when the differences between seen environments are too small to reliably identify the invariant mechanism or the unseen environment requires extrapolation beyond the training data boundaries.
Models without gating (Flow, ANM) seem to be slightly more robust in this respect.
A detailed analysis of our experiments can be found in Appendix \ref{app:experimentalSetting}.


\begin{table*}
\caption{Medians and upper $95\%$ quantiles for domain generalization L2 errors (i.e. on unseen environments) for different model types and data-generating mechanisms (lower is better).}
\label{tab:oodMechanisms}
\begin{center}
\begin{small}
\vskip 0.15in
\begin{tabular}{ p{4.4cm} p{1.9cm} p{1.9cm} p{1.9cm} p{1.9cm} p{1.9cm} } 
\toprule
    {Models} & {Linear} & {Tanhshrink} & {Softplus} & {ReLU} & {Mult. Noise} \\ \midrule
FlowG (ours) & $ 1.05 $\scriptsize$... 4.2 $ & $ 1.08 $\scriptsize$... 4.8 $ & $ 1.09 $\scriptsize$... 5.52 $ & $ 1.08 $\scriptsize$... 5.7 $ & $ 1.55 $\scriptsize$... 8.64 $ \\
ANMG (ours) & $ 1.02 $\scriptsize$... 1.56 $ & $ \mathbf{ 1.03 } $\scriptsize$... 2.23 $ & $ \mathbf{ 1.04 } $\scriptsize$... 4.66 $ & $ \mathbf{ 1.03 } $\scriptsize$... 4.32 $ & $ 1.46 $\scriptsize$... 4.22 $ \\
Flow (ours) & $ 1.08 $\scriptsize$... 1.61 $ & $ 1.14 $\scriptsize$... 1.57 $ & $ 1.14 $\scriptsize$... 1.55 $ & $ 1.14 $\scriptsize$... 1.54 $ & $ \mathbf{ 1.35 } $\scriptsize$... 4.07 $ \\
ANM (ours) & $ 1.05 $\scriptsize$... 1.52 $ & $ 1.15 $\scriptsize$... 1.47 $ & $ 1.14 $\scriptsize$... 1.47 $ & $ 1.15 $\scriptsize$... 1.54 $ & $ 1.48 $\scriptsize$... 4.19 $ \\
ICP (Peters et al., 2016) & $ \mathbf{ 0.99 } $\scriptsize$... 25.7 $ & $ 1.44 $\scriptsize$... 20.39 $ & $ 3.9 $\scriptsize$... 23.77 $ & $ 4.37 $\scriptsize$... 23.49 $ & $ 8.94 $\scriptsize$... 33.49 $ \\
ERM & $ 1.79 $\scriptsize$... 3.84 $ & $ 1.89 $\scriptsize$... 3.89 $ & $ 1.99 $\scriptsize$... 3.71 $ & $ 2.01 $\scriptsize$... 3.62 $ & $ 2.08 $\scriptsize$... 5.86 $ \\ \midrule
CERM (true parents) & $ 1.06 $\scriptsize$... 1.89 $ & $ 1.06 $\scriptsize$... 1.84 $ & $ 1.06 $\scriptsize$... 2.11 $ & $ 1.07 $\scriptsize$... 2.15 $ & $ 1.37 $\scriptsize$... 5.1 $ \\
\end{tabular}

\end{small}
\end{center}
\vskip -0.2in
\end{table*}

\subsection{Colored MNIST}
\label{sec:colored-mnist}

\begin{figure}[!b]
\vspace{-2mm}
\centering
\includegraphics[trim={12 0 0 0},clip,width=0.8\linewidth]{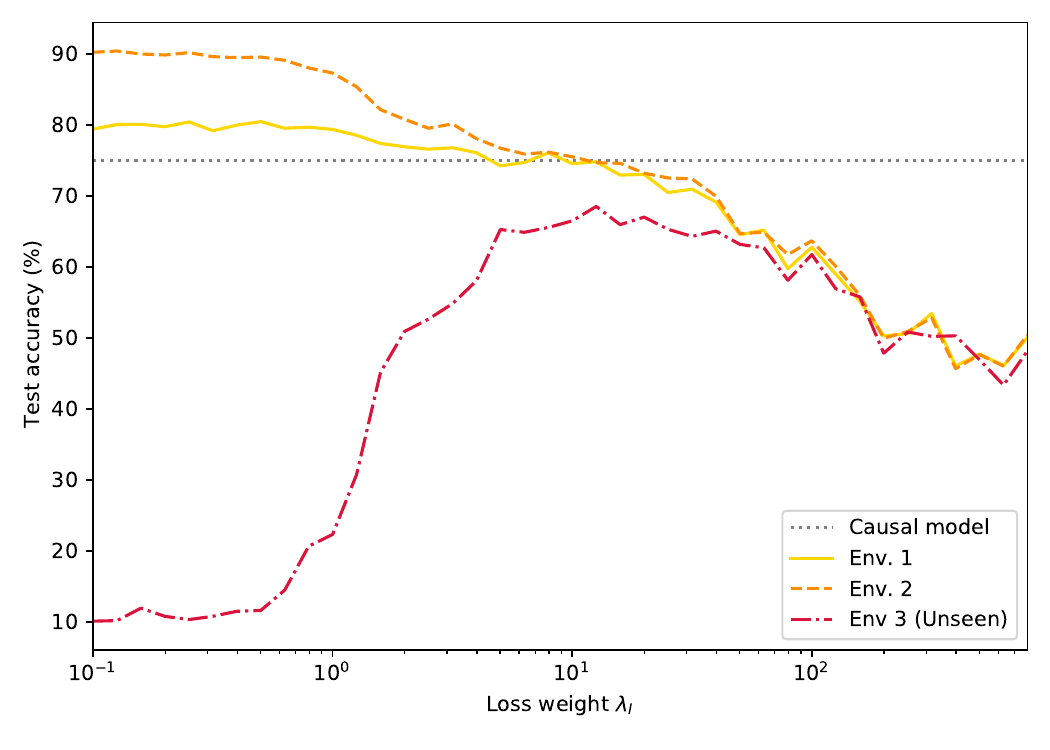}
\captionof{figure}{
Performance of the model in the three environments, depending on the hyperparameter $\lambda_I$.}
\label{fig:mnist_lambda}
\end{figure}

To demonstrate that our model is able to perform DG in the classification case, we use the same data generating process as in the colored variant of the MNIST-dataset established by \citet{arjovsky2019invariant}, but create training instances online rather than upfront.
The response is reduced to two labels -- $0$ for all images with digit $\{0,\dots, 4\}$ and $1$ for digits $\{5, \dots 9\}$ -- with deliberate label noise that limits the achievable shape-based classification accuracy to 75\%.
To confuse the classifier, digits are additionally colored such that colors are spuriously associated with the true labels at accuracies of 90\% resp. 80\% in the first two environments, whereas the association is only 10\% correct in the third environment.
A classifier naively trained on the first two environments will identify color as the best predictor, but will perform terribly when tested on the third environment.
In contrast, a robust model will ignore the unstable relation between colors and labels and use the invariant relation, namely the one between digit shapes and labels, for prediction. 
We supplement the HSIC loss with a Wasserstein term to explicitly enforce $R \perp E$, i.e. $\mathcal{L}_{I} = \mathrm{HSIC} + \mathrm{L2}( \mathrm{sort}(R^{e_1}), \mathrm{sort}(R^{e_2}))$ (see Appendix \ref{app:HSIC}).
This gives a better training signal as the HSIC alone, since the difference in label-color association between environments 1 and 2 (90\% vs. 80\%) is deliberately chosen very small to make the task hard to learn. Experimental details can be found in Appendix \ref{app:colorMnistExDet}.
Figure \ref{fig:mnist_bar_graph} shows the results for our model:
Naive training ($\lambda_I=0$, i.e. invariance of residuals is not enforced) gives accuracies corresponding to the association between colors and labels and thus completely fails in test environment 3. 
In contrast, our model performs close to the  best possible rate for invariant classifiers in environments 1 and 2 and still achieves 68.5\% in environment 3.
Figure \ref{fig:mnist_lambda} demonstrates the trade-off between goodness of fit in the training environments 1 and 2 and the robustness of the resulting classifier: the model's ability to perform DG to the unseen environment 3 improves as $\lambda_I$ increases. If $\lambda_I$ is too large, it dominates the classification training signal and performance breaks down in all environments. However, the choice of $\lambda_I$ is not critical, as good results are obtained over a wide range of settings.

\begin{figure}[t]

\includegraphics[trim={0 0 30 0},clip,width=0.8\linewidth]{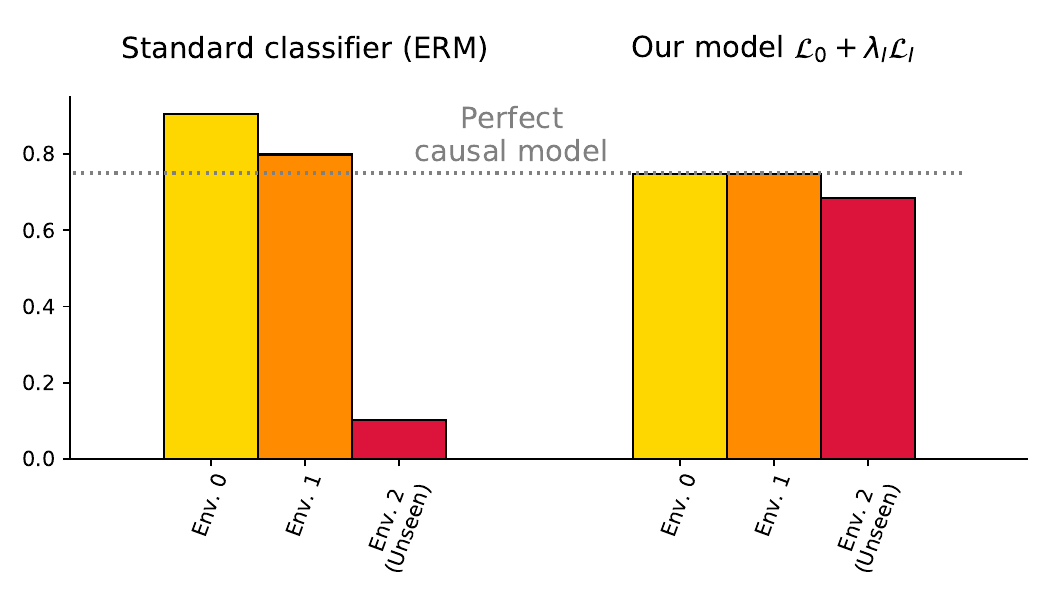}
\centering
\begin{tabular}{l | r r r}
  &   Env. 1 & Env. 2 & Env. 3 \\ \hline
 ERM & 90.3 &  79.9 &  10.2 \\
$\IL_0 + \lambda_I \IL_I $ & 74.8 &  74.7 &  68.5
\end{tabular}
\captionof{figure}{
Accuracy of a standard classifier
and our model}
\label{fig:mnist_bar_graph}
\vspace{-2mm}
\end{figure}

\section{Conclusions}
\vspace{-2mm}
In this paper, we have introduced a new method to find invariant and causal models by exploiting the principle of ICM. Our method works by gradient descent in contrast to combinatorial optimization procedures. 
This circumvents scalability issues and allows us to extract invariant features even when the raw data representation is not in itself meaningful (e.g.~we only observe pixel values). 
In comparison to alternative approaches, our use of normalizing flows places fewer restrictions on the underlying true generative process.
We have also shown under which circumstances our method guarantees to find the underlying causal model. 
Moreover, we demonstrated theoretically and empirically that our method is able to learn robust models w.r.t. distribution shift. 
As a next step, we will examine our approach in more complex scenarios where, for instance, the invariance assumption may only hold approximately.
%\section{Discussion}
%\input{07_discussion.tex}
%\subsubsection*{Author Contributions}
%If you'd like to, you may include  a section for author contributions as is done
%in many journals. This is optional and at the discretion of the authors.

\subsubsection*{Acknowledgments}

JM received funding by the Heidelberg Collaboratory for Image Processing (HCI). 
LA received funding by the Federal Ministry of Education and Research of Germany project High Performance Deep Learning Framework (No 01\,IH\,17002). CR and UK received financial support from the European Research Council (ERC) under the European Unions Horizon2020 research and innovation program (grant agreement No647769).

Furthermore, we thank our colleagues Felix Draxler, Jakob Kruse and Michael Aichmüller for their help, support and fruitful discussions.

\FloatBarrier

\bibliography{bibliography}
\bibliographystyle{icml2021}

\newpage
\begin{appendix}
\section*{Appendix}
\label{sec:Appendix}

\section{Causality: Basics}
\label{app:causalbasics}

%%%%%%% Description SCM
\emph{Structural Causal Models} (SCM) allow us to express causal relations on a functional level. Following \citet{peters2017elements} we define a SCM in the following way:

\begin{definition}
    \label{def:SCM}
	A Structural Causal Model (SCM) $\mathcal{S}=(S ,P_{\pmb{N}})$ consists of a collection $S$ of $D$ (structural) assignments
    \begin{align}
        \label{eq:scmDefinition}
        X_j \coloneqq f_j(\WX_{pa(j)}, N_j), \ \ \ j=1, \dots,D
    \end{align}
    where $pa(j) \subset \{1, \dots , j-1 \}$ are called parents of $X_j$. $P_{\pmb{N}}$ denotes the distribution over the noise variables $\pmb{N} = (N_1, \dots, N_D)$ which are assumed to be jointly independent.
\end{definition}
An SCM defined as above produces an acyclic graph $G$ and induces a probability distribution over $P_{\widetilde{\X}}$ which allows for the \emph{causal factorization} as in \eqref{eq:interventionalFactorization}
\citet{peters2017elements}. 
Children of $X_i$ in $G$ are denoted as $ch(i)$ or $ch(X_i)$. 
An SCM satisfies the \emph{causal sufficiency} assumption if all the noise variables in Definition \ref{def:SCM} are indeed jointly independent. A random variable $H$ in the SCM is called \emph{confounder} between two variables $X_i, X_j$ if it causes both of them. If a confounder is not observed, we call it hidden confounder. If there exists a hidden confounder, the causal sufficiency assumption is violated.

The random variables in an SCM correspond to vertices in a graph and the structural assignments $S$ define the edges of this graph. 
Two sets of vertices $\pmb{A}, \pmb{B}$ are said to be $d$-separated if there exists a set of vertices $\pmb{C}$ such that every path between $\pmb{A}$ and $\pmb{B}$ is blocked. For details see e.g. \citet{peters2017elements}.  The subscript $\perp_d$ denotes $d$-separability which in this case is denoted by $\pmb{A} \perp_d \pmb{B}$. An SCM generates a probability distribution $P_{\widetilde{\X}}$ which satisfies the \emph{Causal Markov Condition}, that is  $\pmb{A} \perp_d \pmb{B} \mid \pmb{C}$ results in $\pmb{A} \perp \pmb{B} \mid \pmb{C}$ for sets or random variables $\pmb{A}, \pmb{B}, \pmb{C} \subset \widetilde{\X}$. The Causal Markov Condition can be seen as an inherent property of a causal system which leaves marks in the data distribution.

%%%% Faithfulness and Causal Markov Condition
A distribution $P_{\widetilde{\X}}$ is said to be \emph{faithful} to the graph $G$ if $\pmb{A} \perp \pmb{B} \mid \pmb{C}$ results in $\A \perp_d \pmb{B} \mid \pmb{C}$ for all $\A, \B, \C \subset \WX$. This means from the distribution $P_{\WX}$ statements about the underlying graph $G$ can be made.

Assuming both, faithfulness and the Causal Markov condition, we obtain that the $d$-separation statements in $G$ are equivalent to the conditional independence statements in $P_{\WX}$. These two assumptions allow for a whole class of causal discovery algorithms like the PC- or IC-algorithm \citep{spirtes1991algorithm, pearl2009causality}.

%%%%%%%% Markov Blanket
The smallest set $\pmb{M}$ such that $ Y \perp_d \X \setminus ( \{Y \} \cup \pmb{M}) $ is called \emph{Markov Blanket}. It is given by $\pmb{M} =  \X_{pa(Y)} \cup \X_{ch(Y)} \cup 
\X_{pa(ch(Y))}\setminus \{Y \}$. The \emph{Markov Blanket} of $Y$ is the only set of vertices which are necessary to predict $Y$. 

\section{Discussion and Illustration of Assumptions}
\label{app:discussAssumptions}

\subsection{Examples}
Domain generalization is in general impossible without strong assumptions (in contrast to classical supervised learning). In our view, the interesting question is “Which strong assumptions are the most useful in a given setting?”.  
For instance, \cite{heinze2018invariant} use Assumption 2 to identify causes for birth rates in different countries. If all variables mediating the influence of continent/country (environment variable) on birth rates (target variable) are included in the model (e.g. GDP, Education), this assumption is reasonable. The same may hold for other epidemiological investigations as well. \cite{pfister2019invariant} suppose Assumption 2 in the field of finance.

Another reasonable example are data augmentations in computer vision. Deliberate image rotations, shifts and distortions can be considered as environment interventions that preserve the relation between semantic image features and object classes (see e.g. \cite{mitrovic2020representation}), i.e. verify assumption 2. In general, assumption 2 may be justified when one studies a fundamental mechanism that can reasonably be assumed to remain invariant across environments, but is obscured by unstable relationships between observable variables.

\subsection{Causal Sufficiency}

Violation of the causal sufficiency assumption might prevent Assumption 2 to hold true. For instance, a causal graph with edges $H\to X_1, X_1 \to Y$ and $H\to Y$ where $H$ is not observed, violates the causal sufficiency assumption. 
If the environment influences $X_1$, i.e. the graph also contains edge $E \to X_1$, and the generated distribution satisfies the Causal Markov Condition, it follows that $Y \perp E \mid X_1$ is unachievable. This example also illustrates also that the causal sufficiency assumption is necessary for the principle of ICM.

\subsection{Robustness}

To illustrate the impact of causality on robustness, consider the following example: 
Suppose we would like to estimate the gas consumption of a car.
In a sufficiently narrow setting, the total amount of money spent on gas might be a simple and accurate predictor.
However, gas prices vary dramatically between countries and over time, so statistical models relying on it will not be robust, even if they fit the training data very well.
Gas costs are an {\em effect} of gas consumption, and this relationship is unstable due to external influences.
In contrast, predictions on the basis of the {\em causes} of gas consumption (e.g. car model, local speed limits and geography, owner's driving habits) tend to be much more robust, because these causal relations are intrinsic to the system and not subjected to external influences.
Note that there is a trade-off here:
Including gas costs in the model will improve estimation accuracy when gas prices remain sufficiently stable, but will impair results otherwise.
By considering the same phenomenon in several environments simultaneously, we hope to gain enough information to adjust this trade-off properly.

In the gas example, countries can be considered as environments that ``intervene'' on the relation between consumed gas and money spent, e.g. by applying different tax policies. In contrast, interventions changing the impact of motor properties or geography on gas consumption are much less plausible – powerful motors and steep roads will always lead to higher consumption. From a causal standpoint, finding robust models is therefore a causal discovery task \citet{meinshausen2018causality}.

\section{Normalizing Flows}
\label{app:normFlows}

Normalizing flows are a specific type of neural network architecture which are by construction invertible and have a tractable Jacobian. They are used for density estimation and sampling of a target density (for an overview see \citet{papamakarios2019normalizing}). This in turn allows optimizing information theoretic objectives in a convenient and mathematically sound way. 

Similarly as in the paper, we denote with $\mathcal{H}$ the set of feature extractors $h \colon \R^D \to \R^M$ where $M$ is chosen a priori. 
The set of all one-dimensional (conditional) normalizing flows is denoted by $\mathcal{T}$.
Together with a reference distribution $p_{ref}$, a normalizing flow $T$ defines a new distribution $\nu_T = (T( \cdot; \h))^{-1}_{\#}p_{ref}$ which is called the push-forward of the reference distribution $p_{ref}$  \citep{marzouk2016sampling}. By drawing samples from $p_{ref}$ and applying $T$ on these samples we obtain samples from this new distribution. The density of this so-obtained distribution $p_{\nu_T}$ can be derived from the change of variables formula:
\begin{align}
    \label{eq:chVariables}
    p_{\nu_T} (y\mid \h) = p_{ref} (T(y; \h))  | \nabla_y T(y; \h)| 
\end{align}
The KL-divergence between the target distribution $p_{Y\mid \HB}$ and the flow-based model $p_{\nu_T}$ can be written as follows:

\begin{align}
    \label{eq:KLDivLoss}
        &\E_{\HB} [ D_{\text{KL}} ( p_{Y\mid \HB} \| p_{\nu_T}) ] \nonumber \\ = &\E_{\HB} \left[\E_{Y \mid \HB}\left[ \log \Big(\frac{ p_{Y \mid \HB}}{ p_{ \nu_T }}  \Big) \right]\right] \nonumber \\
        = & -H( Y \mid \HB ) - \E_{h(\X),Y} [\log p_{\nu_T} (Y\mid  \HB) ] \nonumber \\
        = & -H(Y \mid \HB) + \E_{\HB, Y} [ -\log p_{ref} (T(y;\h) \nonumber \\ &- \log| \nabla_y T(y; \h)| ]
\end{align}
The last two terms in \eqref{eq:KLDivLoss} correspond to the negative log-likelihood (NLL) for conditional flows with distribution $p_{ref}$ in latent space. 
If the reference distribution is assumed to be standard normal, the NLL is given as in Section \ref{sec:preliminaries}.

We restate Lemma \ref{lem:normalizingFlows} with a more general notation. Note that the argmax or argmin is a set.
\begin{repeatlemma}{lem:normalizingFlows}
Let $\X, Y$ be random variables. We furthermore assume that for each $h \in \mathcal{H}$ there exists one $T \in \mathcal{T}$ with $\E_{\HB} [ D_{\text{KL}} ( p_{Y\mid \HB} \| p_{\nu_T}) ] = 0$. Then, the following two statements are true
\begin{itemize}
    \item[(a)] Let
    \begin{align*}
        h^{\star}, T^{\star} = \arg \min_{h \in \mathcal{H}, T \in \mathcal{T}}  - \E_{h(\X),Y} [\log p_{\nu_T} (Y\mid  h(\X)) ]
    \end{align*}
    then it holds $h^{\star} = g^{\star}$ where $g^\star =  \arg \max_{g \in \mathcal{H}} I(g(\X); Y)$
    \item[(b)] Let
    \begin{align*}
            T^{\star} = \arg \min_{T\in \mathcal{T}} \E_{h(\X)} [ D_{\text{KL}} ( p_{Y\mid h(\X)} \| p_{\nu_T}) ] 
    \end{align*}
    then it holds $h(\X) \perp T^{\star}(Y; h(\X)) $
\end{itemize}
\end{repeatlemma}

\begin{proof}
(a) From \eqref{eq:KLDivLoss}, we obtain 
$- \E_{h(\X),Y} [\log p_{\nu_T} (Y\mid  h(\X)) ] \ge  H(Y \mid h(\X))$ 
for all 
$h\in \mathcal{H}, T \in \mathcal{T}$. 
We furthermore have 
$\min_{T \in \mathcal{T}} - \E_{h(\X),Y} [\log p_{\nu_T} (Y\mid  h(\X)) ] =  H(Y\mid h(\X))$ 
due to our assumptions on $\mathcal{T}$. 
Therefore,  
$\min_{h \in \mathcal{H}, T \in \mathcal{T}}  - \E_{h(\X),Y} [\log p_{\nu_T} (Y\mid  h(\X)) ]  =  \min_{h \in \mathcal{H}} H(Y \mid h(\X))$. 
Since we have $I(Y; h(\X)) = H(Y) - H(Y \mid h(\X))$ and only the second term depends on $h$, statement (a) holds true.

(b) 
For convenience, we denote $T(Y; h(\X)) = R$ and $h(\X) = Z $. 
We have $\E_{Z} [ D_{\text{KL}} ( p_{Y\mid Z} \| p_{\nu_{T^\star}})] = 0$ and therefore $p_{Y\mid Z} (y \mid z) = p_{ref} (T(y;z)) | \nabla_y T^{-1}(y; z)| $.

Then it holds 
\begin{align*}
 p_{R \mid Z} (r \mid z) &= p_{Y\mid Z}(T^{-1}(r;z)| z) \cdot | \nabla_y T^{-1}(r; z)| \\
 &= p_{ref} (T (T^{-1} (r;z);z)) \cdot | \nabla_y T(y; z)| \\ & \quad \cdot | \nabla_y T^{-1}(r; z)| \\
 &= p_{ref} (r) \cdot 1
\end{align*}
Since the density $p_{ref}$ is independent of $Z$, we obtain $R \perp Z$ which concludes the proof of (b)
\end{proof}

Statement (a) describes an optimization problem that allows to find features which share maximal information with the target variable $Y$. Due to statement (b) it is possible to draw samples from the conditional distribution $P(Y \mid h(\X))$ via the reference distribution.

Let $\mathcal{H}_\perp$ the set of features which satisfy the invariance property, i.e.~$Y \perp E \mid h(\X)$ for all $h \in \mathcal{H}_\perp$. 
In the following, we sketch why $\argmin_{h \in \mathcal{H}_\perp } \max_{e \in \IE} \min_{T \in \mathcal{T}} \mathcal{L}_\mathrm{NLL}^e(T;h) = \argmax_{h \in \mathcal{H}_\perp} \min_{e \in \IE} I(Y^e; h(\X^e))$ follows from Lemma \ref{lem:normalizingFlows}.

Let $h \in \mathcal{H}_\perp$. Then, it is easily seen that there exists a $T^\star \in \mathcal{T}$ with (1) $\mathcal{L}_{\mathrm{NLL}}(T^\star;h) = \min_{T \in \mathcal{T}} \mathcal{L}_{\mathrm{NLL}} (T, h)$ and (2) $\mathcal{L}_{\mathrm{NLL}}^e(T^\star,h) = \min_{T \in \mathcal{T}} \mathcal{L}_\mathrm{NLL}^e (T, h)$ for all $e \in \IE$ since the conditional densities $p(y\mid h(\X))$ are invariant across all environments.
Hence we have $H(Y^e \mid h(\X^e)) =  \mathcal{L}_\mathrm{NLL}^e (T^\star;h)$ for all $e \in \IE$. Therefore, 
$\argmin_{h \in \mathcal{H}_\perp } \max_{e \in \IE} \min_{T \in \mathcal{T}} \mathcal{L}_\mathrm{NLL}^e(T;h) = \argmax_{h \in \mathcal{H}_\perp} \min_{e \in \IE} I(Y^e; h(\X^e))$
due to $I(Y^e; h(\X^e)) = H(Y^e) - H(Y^e \mid h(\X^e))$.

\section{HSIC and Wasserstein}
\label{app:HSIC}

The Hilbert-Schmidt Independence Criterion (HSIC) is a kernel based measure for independence which is in expectation $0$ if and only if the compared random variables are independent \citep{gretton2005measuring}. An empirical estimate of $\text{HSIC}(A, B)$ for two random variables $A, B$ is given by 
\begin{align}
    \widehat{\text{HSIC}} ( \{a_j\}_{j=1}^n, \{b_j\}_{j=1}^n ) = \frac{1}{(n-1)^2} \operatorname{tr}(KHK'H)
\end{align}
where $\operatorname{tr}$ is the trace operator. $K_{ij}= k(a^i,a^j)$ and $K'_{ij} =k'(b^i,b^j)$ are kernel matrices for given kernels $k$ and $k'$. The matrix $H$ is a centering matrix $H_{i,j} = \delta_{i,j} - 1/n$.

The one dimensional Wasserstein loss compares the similarity of two distributions \citep{kolouri2018sliced}. This loss has expectation $0$ if both distributions are equal. An empirical estimate of the one dimensional Wasserstein loss for two random variables $A,B$ is given by
\begin{align*}
\mathcal{L}_W= \|\mathrm{sort}( \{a_j\}_{j=1}^n)-  \mathrm{sort}(\{b_j\}_{j=1}^n )\|_2
\end{align*} 
Here, the two batches are sorted in ascending order and then compared in the L2-Norm. We assume that both batches have the same size.
\begin{figure}
    \centering
      \includegraphics[width=.8\linewidth]{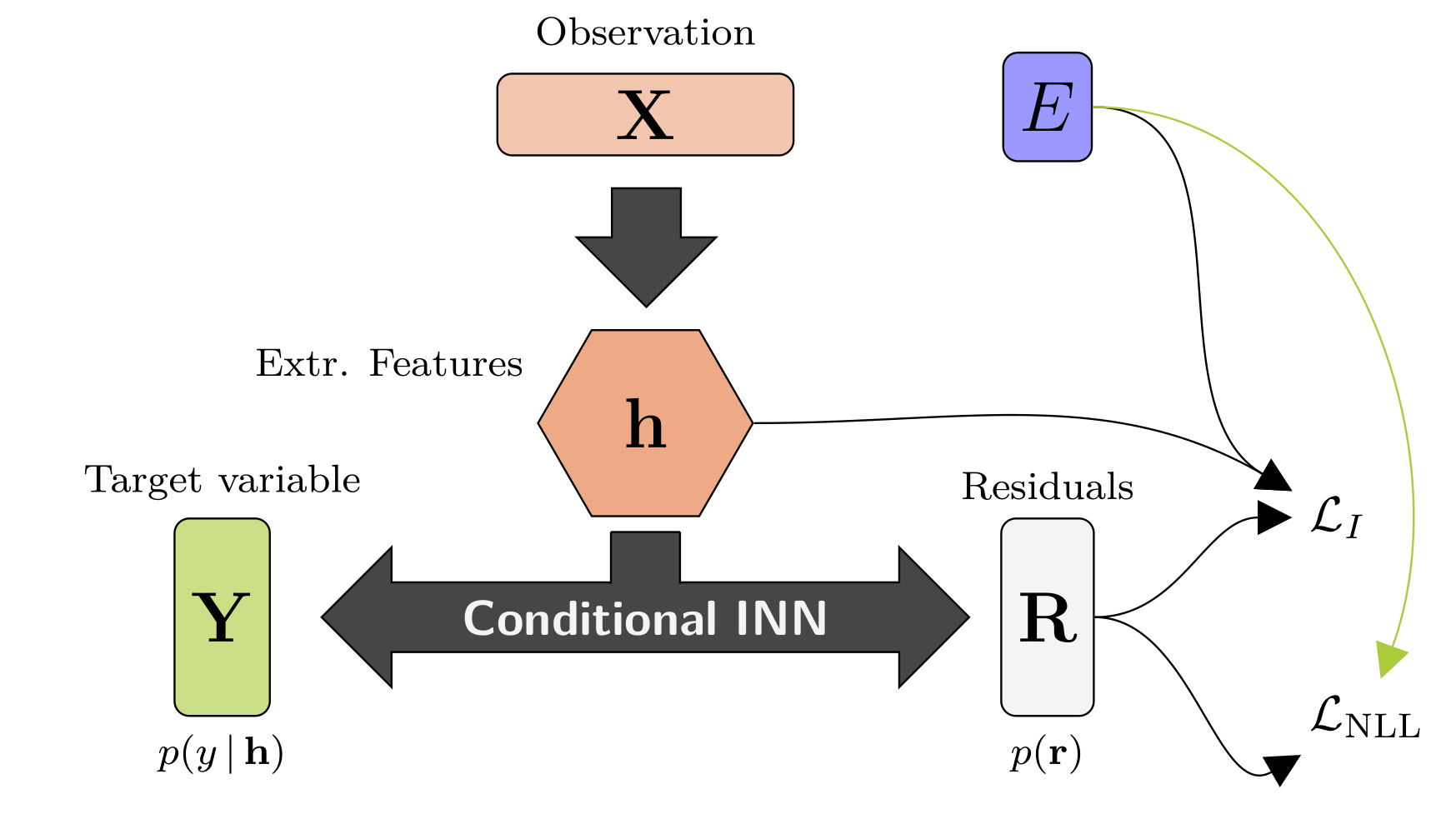}  
    \caption{Illustration of Architecture of normalizing flow model which implements \eqref{eq:flow-and-hsic-loss}. $h$ is a feature extractor implemented as feed forward neural network.}
    \label{fig:architectureFlow}
\end{figure}

\section{Algorithm}
\label{app:algo}

In order to optimize the DG problem in \eqref{eq:ourEstimator2}, we optimize a normalizing flow $T_\theta$ and a feed forward neural network $h_\phi$ as described in Algorithm \ref{alg:residual}. There is an inherent trade-off between robustness and goodness-of-fit. The hyperparameter $\lambda_I$ describes this trade-off and is chosen a priori.

\begin{algorithm}
 \KwData{Samples from $P_{\X^e, Y^e}$ in different environments $e \in \IES$.}
 \SetKwInOut{Init}{Initialize}
\Init{Parameters $\theta, \phi$;}
 \For{number of training iterations}{
 \For{$e \in \IES$}{
    Sample minibatch $\{ (y_1^e, \x_1^e), \dots, (y_m^e, \x_m^e) \}$ from $P_{Y, \X \mid E=e}$ for $e \in \IES$;\;
    
    Compute $r^e_j = T_\theta ( y_j^e; h_\phi (\x_j^e))$;\;
    }
    Update $\theta, \phi$ by descending alongside the stochastic gradient
        \begin{align*}
              \nabla_{\theta, \phi}   \Big(& \max_{e \in \IES} \Big\{ \sum_{i=1}^{m} \big[\tfrac{1}{2}  \|  T_\theta (y^e_i; h_\phi(\x^e_i)) \|^2 \nonumber  \\ &- \log \nabla_y T_\theta(y^e_i; h_\phi(\x^e_i)) \big] \Big\} \nonumber  \\
              &+ \lambda_I \IL_I (\{r^e_j\}_{j,e}, \{ h_\phi(\x^e_j), e \}_{j,e}) \Big);\;
        \end{align*} 
 }
 \KwResult{In case of convergence, we obtain $ T_{\theta^\star}, h_{\phi^\star}$ with 
 \begin{align*}
 \theta^\star, \phi^\star = & \\ \arg \min_{\theta, \phi}   \Big( & \max_{e \in \IES} \Big\{ \E_{\X^e,Y^e}\big[\tfrac12 \|  T_\theta (Y^e; h_\phi(\X^e)) \|^2 \nonumber \\
 & - \log \nabla_y T_\theta(Y^e; h_\phi(\X^e))  \big] \Big \} \nonumber \\ & + \lambda_I \IL_I (P_R, P_{h_\phi(\X), E}) \Big)
 \end{align*}}
 \caption{DG training with normalizing flows}
 \label{alg:residual}
\end{algorithm}

\begin{figure}[t]
    \centering
      \includegraphics[width=.8\linewidth]{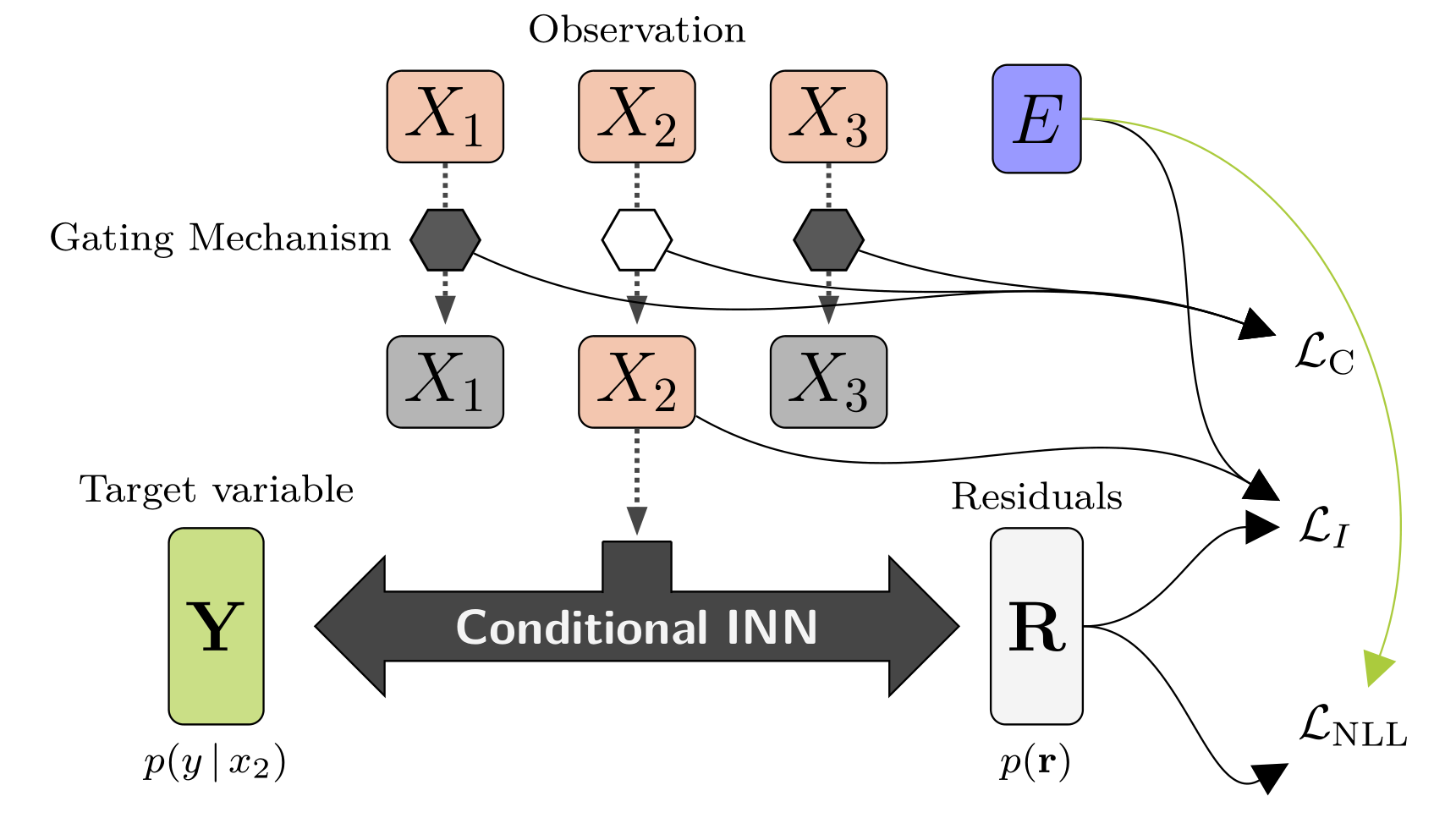}  
    \caption{Illustration of Architecture of normalizing flow model which implements \eqref{eq:flow-and-hsic-loss} where the feature extractor $h$ is a gating mechanism. Architecture is depicted for three input variables.}
    \label{fig:architectureFlowGating}
\end{figure}

If we choose a gating mechanisms $h_\phi$ as feature extractor similar to \citet{kalainathan2018sam}, then a complexity loss is added to the loss in the gradient update step. The architecture is illustrated in Figure \ref{fig:architectureFlow}. Figure \ref{fig:architectureFlowGating} shows the architecture with gating function. 

In case we assume that the underlying mechanisms elaborates the noise in an additive manner, we could replace the normalizing flow $T_\theta$ with a feed forward neural network $f_\theta$ and execute Algorithm \ref{alg:residualsAddNoise}.

\begin{algorithm}%[tb]%[H]
 \caption{DG training under the assumption of additive noise}
 \label{alg:residualsAddNoise}
 \KwData{Samples from $P_{\X^e, Y^e}$ in different environments $e \in \IES$.}
 %Initialize $\theta$\;
 \SetKwInOut{Init}{Initialize}
\Init{Parameters $\theta, \phi$;}
 \For{number of training iterations}{
 \For{$e \in \IES$}{
    Sample minibatch $\{ (y_1^e, \x_1^e), \dots, (y_m^e, \x_m^e) \}$ from $P_{Y, \X \mid E=e}$ for $e \in \IES$;\;
    
    Compute $r^e_j = y_j^e - f_\theta( \x_j^e) $;\;

    }
    Update $\theta$ by descending alongside the stochastic gradient
    \begin{align*}
          \nabla_{\theta}   \Big( & \max_{e \in \IES} \Big\{ \sum_{i=1}^{m} | r_j^e | ^2  \Big\} 
          \nonumber \\ &+ \lambda_I \IL_I (\{r^e_j\}_{j,e}, \{ f_\theta(\x^e_j), e \}_{j,e}) \Big);\;
    \end{align*} 
 }
 \KwResult{In case of convergence, we obtain $f_{\theta^\star}$ with 
 \begin{align*}
 \theta^\star = \arg \min_{\theta}   \Big( & \max_{e \in \IES} \Big\{ \E_{\X^e,Y^e}\big[|Y^e- f_\theta(\X^e)|^2 \big] \nonumber \Big \} \nonumber \\
 & + \lambda_I \IL_I (P_R, P_{f_\theta(\X), E}) \Big)
 \end{align*}}

\end{algorithm}

If we choose a gating mechanism, minor adjustments have to be made to Algorithm \ref{alg:residualsAddNoise} such that we optimize \eqref{eq:additiveNoiseDG2}. The classification case can be obtained similarly as described in Section \ref{sec:methods}.

\section{Identifiability Result}
\label{app:identifiability}

Under certain conditions on the environment and the underlying causal graph, the direct causes of $Y$ become identifiable:

\begin{repeatprop}{prop:identifiability2}
	 We assume that the underlying causal graph $G$ is faithful with respect to $P_{\WX,  E}$.
	We further assume that %$ch(Y) \cap ch(E) = ch(Y)$, i.e. 
	every child of $Y$ in $G$ is also a child of $E$ in $G$. A variable selection $h(\X)= \X_S$ corresponds to the direct causes if the following conditions are met: (i) $T(Y;h(\X)) \perp E, h(\X)$ are satisfied for a diffeomorphism $T(\cdot ; h(\X))$, (ii) $h(\X)$ is maximally informative about $Y$ and (iii) $h(\X)$ contains only variables from the Markov blanket of $Y$.
\end{repeatprop}
\begin{proof}
    Let $S(\IES)$ denote a subset of $\X$ which corresponds to the variable selection due to $h$. Without loss of generality, we assume $S(\IES) \subset \M$ where $\M$ is the Markov Blanket. This assumption is reasonable since we have $Y \perp \X \setminus  \M  \mid \M $ in the asymptotic limit.

    Since $pa(Y)$ cannot contain colliders between $Y$ and $E$, we obtain that $Y \perp E \mid S(\IES)$ implies $Y \perp E \mid (S(\IES)\cup pa(Y)) $. This means using $pa(Y)$ as predictors does not harm the constraint in the optimization problem. Due to faithfulness and since the parents of $Y$ are directly connected to $Y$, we obtain that $pa(Y) \subset S(\IES)$.

	For each subset $\X_S \subset \X$ for which there exists an $X_i \in \X_S \cap \X_{ch(Y)}$, we have $\X_S \not \perp Y \mid E$. This follows from the fact that $X_i$ is a collider, in particular $E \to X_i \leftarrow Y$. Conditioning on $X_i$ leads to the result that $Y$ and $E$ are not $d$-separated anymore. Hence, we obtain $Y \not \perp \X_S \mid E$ due to the faithfulness assumption. 
	Hence, for each $\X_S$ with $Y \perp E \mid \X_S$ we have $\X_S \cap \X_{ch(Y)} = \emptyset$ and therefore $\X_{ch(Y)} \cap S(\IES) = \emptyset $.

    Since $\X_{pa(Y)} \subset S(\IES)$, we obtain that $Y \perp \X_{pa(ch(Y))} \mid \X_{pa(Y)}$ and therefore the parents of $ch(Y)$ are not in $S(\IES)$ except when they are parents of $Y$. 
    
    Therefore, we obtain that $S(\IES) = \X_{pa(Y)}$
	
\end{proof}

One might argue that the conditions are very strict in order to obtain the true direct causes. But the conditions set in Proposition \ref{prop:identifiability2} are necessary if we do not impose additional constraints on the true underlying causal mechanisms, e.g. linearity as done by \citet{peters2016causal}. For instance if $E \to X_1 \to Y \to X_2$, a model including $X_1$ and $X_2$ as predictor might be a better predictor than the one using only $X_1$. From the Causal Markov Condition we obtain $E \perp Y \mid X_1, X_2$ which results in $X_1, X_2 \in S(\IES)$. Under certain conditions however, the relation $Y \to X_2$ might be invariant across $\IE$. This is for instance the case when $X_2$ is a measurement of $Y$. In this cases it might be useful to use $X_2$ for a good prediction.

\section{Experimental Setting for Synthetic Dataset}
\label{app:experimentalSetting}

\subsection{Data Generation}
\label{app:dataGeneration}

In Section \ref{sec:experiments} we described how we choose different Structural Causal Models (SCM). In the following we describe details of this process.

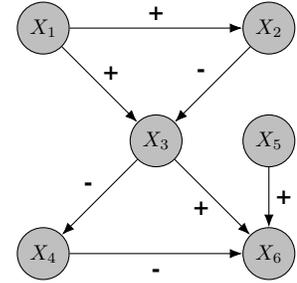
\begin{wrapfigure}{r}{0.25\textwidth}
\centering
\begin{tikzpicture}
	\node[state, circle, scale=0.8, fill=gray!50] (x0) at (-1.5, 1.5) {$X_1$};
	\node[state, circle, scale=0.8, fill=gray!50] (x1) at (1.5,1.5) {$X_2$};
	\node[state, circle, scale=0.8, fill=gray!50] (x2) at (0, 0) {$X_3$};
	\node[state, circle, scale=0.8, fill=gray!50] (x3) at (-1.5, -1.5) {$X_4$};
	\node[state, circle, scale=0.8, fill=gray!50] (x4) at (1.5, 0) {$X_5$};
	
	\node[state, circle, scale=0.8, fill=gray!50] (x5) at (1.5, -1.5) {$X_6$};
    \node[] at (0,1.7) {\textbf{+}};
    \node[] at (-0.6,0.9) {\textbf{+}};
    \node[] at (0.6,0.9) {\textbf{-}};
    \node[] at (-0.9,-0.6){\textbf{-}};
    \node[] at (0.6,-0.9) {\textbf{+}};
    \node[] at (1.7,-0.75) {\textbf{+}};
    \node[] at (0,-1.75) {\textbf{-}};

	\path (x0) edge (x2);
	\path (x0) edge (x1);
	\path (x1) edge (x2);
	\path (x2) edge (x3);
	\path (x2) edge (x5);
	\path (x4) edge (x5);
	\path (x3) edge (x5);
\end{tikzpicture}
    \caption{The signs of the coefficients $a_j$ for the mechanisms of the different SCMs}
    \label{fig:heinze_graphSigns}
\end{wrapfigure}

We simulate the datasets in a way that the conditions in Proposition \ref{prop:identifiability2} are met. We choose different variables in the graph shown in Figure \ref{fig:heinze_graph} as target variable. Hence, we consider different ``topological'' scenarios. We assume the data is generated by some underlying SCM. We define the structural assignments in the SCM as follows 
\begin{alignat*}{2}
    &  (\text{a}) \quad f_{i}^{(1)} ( \X_{pa(i)}, N_i) = \sum_{j \in pa(i)} a_j X_j +N_i  \quad \text{[Linear]} & \\ 
    & (\text{b}) \quad f_{i}^{(2)} ( \X_{pa(i)}, N_i) = \sum_{j \in pa(i)} a_j X_j -  \tanh (a_j X_j )+ N_i & \\ & \quad \quad \text{[Tanhshrink]}  & \\
    & (\text{c}) \quad f_{i}^{(3)} ( \X_{pa(i)}, N_i) = \sum_{j \in pa(i)} \log(1+ \exp( a_j X_j)) +N_i  &\\ & \quad \quad \text{[Softplus]}  & \\
    & (\text{d}) \quad f_{i}^{(4)} ( \X_{pa(i)}, N_i) = \sum_{j \in pa(i)} \max \{ 0,  a_j X_j) \} +N_i  & \\ & \quad \quad \text{[ReLU]} & \\
    & (\text{e}) \quad f_{i}^{(5)} ( \X_{pa(i)}, N_i) = \Big(\sum_{j \in pa(i)} a_j X_j \Big) \cdot ( 1+ \frac{1}{4} N_i) +N_i  & \\ & \quad \quad \text{[Mult. Noise]} & \\
\end{alignat*}
 with $N_i \sim \mathcal{N}(0,c_i^2)$ where $c_i \sim \mathcal{U}[0.8, 1.2]$, $i \in \{0, \dots, 5 \}$ and $a_i \in \{-1,1\}$ according to Figure \ref{fig:heinze_graphSigns}.
Note that the mechanisms in (b), (c) and (d) are non-linear with additive noise and (e) elaborates the noise in a non-linear manner.

We consider hard- and soft-interventions on  the assignments $f_i$. We either intervene on all variables except the target variable at once \emph{or} on all parents and children of the target variable (Intervention Location). We consider three types of interventions:
\begin{itemize}
    \item \emph{Hard-Intervention} on $X_i$: Force $X_i \sim e_1+ e_2 \mathcal{N}(0,1)$ where we sample for each environment $e_2 \sim \mathcal{U}([1.5, 2.5])$ and $e_1 \sim \mathcal{U}([0.5,1.5]\cup [-1.5, -0.5])$
    \item \emph{Soft-Intervention} I on $X_i$: Add $e_1+ e_2 \mathcal{N}(0,1)$ to $X_i$ where we sample for each environment $e_2 \sim \mathcal{U}([1.5, 2.5])$ and $e_1 \sim \mathcal{U}([0.5,1.5]\cup [-1.5, -0.5])$

    \item \emph{Soft-Intervention} II on $X_i$: Set the noise distribution $N_i$ to  $\mathcal{N}(0, 2^2)$ for $E=2$ and to $ \mathcal{N}(0,0.2^2)$ for $E=3$
\end{itemize}

Per run, we consider one environment without intervention ($E=1$) and two environments with either both soft- or hard-interventions ($E=2,3$). We also create a fourth environment to measure a models' ability for out-of-distribution generalization:
\begin{itemize}
    \item \emph{Hard-Intervention}: Force $X_i \sim e + \mathcal{N}(0,4^2)$ where $e = e_1 \pm 1$ with $e_1$ from environment $E=1$. The sign $\{+, - \}$ is chosen once for each $i$ with equal probability.
    \item \emph{Soft-Intervention} I: 
    Add $e + \mathcal{N}(0,4^2)$ to $X_i$ where $e = e_1 \pm 1$ with $e_1$ from environment $E=1$. The sign $\{+, - \}$ is chosen once for each $i$ with equal probability as for the \emph{do-intervention} case.
    \item \emph{Soft-Intervention} II: Half of the samples have noise $N_i$ distributed due to $\mathcal{N}(0,1.2^2)$  and the other half of the samples have noise distributed as $\mathcal{N}(0,3^2)$
\end{itemize}

We randomly sample causal graphs as described above. Per environment, we consider $1024$ samples.

\subsection{Training Details}
\label{app:training}

All used feed forward neural networks have two internal layers of size $256$. For the normalizing flows we use a $2$ layer \textit{MTA-Flow} described in Appendix \ref{app:oneDimNormFlows} with K=$32$. As optimizer we use Adam with a learning rate of $10^{-3}$ and a L2-Regularizer weighted by $10^{-5}$ for all models. Each model is trained with a batch size of $256$. We train each model for $1000$ epochs and decay the learning rate every 400 epochs by 0.5. For each model we use $\lambda_I =256$ and the HSIC $\mathcal{L}_I$ employs a Gaussian kernel with $\sigma =1 $. The gating architecture was trained without the complexity loss for $200$ epochs and then with complexity loss weighted by $5$. For the Flow model without gating architecture we use a feed forward neural network $h_\phi$ with two internal layers of size $256$ mapping to an one dimensional vector. In total, we evaluated our models on $1365$ created datasets as described in \ref{app:dataGeneration}.

Once the normalizing flow $T$ is learned, we predict $y$ given features $h(\x)$ using $512$ normally distributed samples $u_i$ which are mapped to samples from $p(y|h(\x))$ by the trained normalizing flow $T(u_i; h(\x))$. As prediction we use the mean of these samples.

\subsection{One-Dimensional Normalizing Flow}
\label{app:oneDimNormFlows}
We use as one-dimension normalizing flow the \textit{More-Than-Affine-Flow (MTA-Flow)}, which was developd by us. An overview of different architectures for one-dimensional normalizing flows can be found in \cite{papamakarios2019normalizing}.
For each layer of the flow, a conditioner network C maps the conditional data $h(\X)$ to a set of parameters $a, b \in \R$ and $\mathbf{w}, \mathbf{v}, \mathbf{r} \in \R^K$ for a chosen $K \in \mathbb{N}$. It builds the transformer $\tau$ for each layer as 
\begin{align}
    z &= \tau(y \mid h(\X)) \nonumber \\ &\coloneqq a \left(y + \frac{1}{\mathop{N}(\mathbf{w}, \mathbf{v})} \sum_{i=1}^K w_i f(v_i y + r_i) \right) + b,
\end{align}
where $f$ is any almost everywhere smooth function with a derivative bounded by 1. In this work we used a gaussian function with normalized derivative for $f$. The division by
\begin{equation}
    N(\mathbf{w}, \mathbf{v}) \coloneqq \varepsilon^{-1} \left(\sum_{i = 1}^K |w_i v_i| + \delta \right),    
\end{equation}
with numeric stabilizers $\varepsilon < 1 $ and $\delta > 0$, assures the strict monotonicity of $\tau$ and thus its invertibility $ \forall x \in \mathbb{R}$. We also used a slightly different version of the \textit{MTA-Flow} which uses the ELU activation function and -- because of its monotonicity -- can use a relaxed normalizing expression $\mathop{N}(\mathbf{w}, \mathbf{v})$.

\subsection{PC-Variant}
\label{app:pcVariant}
Since we are interested in the direct causes of $Y$, the widely applied PC-Algorithm gives not the complete answer to the query for the parents of $Y$. This is due to the fact that it is not able to orient all edges. To compare the PC-Algorithm we include the environment as system-intern variable and use a conservative assignment scheme where non-oriented edges are thrown away. This assignment scheme corresponds to the conservative nature of the ICP. 

For further interest going beyond this work, we consider diverse variants of the PC-Algorithm. We consider two orientation schemes: A \emph{conservative} one, where non-oriented edges are thrown away and a \emph{non-conservative} one where non-oriented edges from a node $X_i$ to $Y$ are considered parents of $Y$. 

\begin{figure}
    \centering
      \includegraphics[width=.8\linewidth]{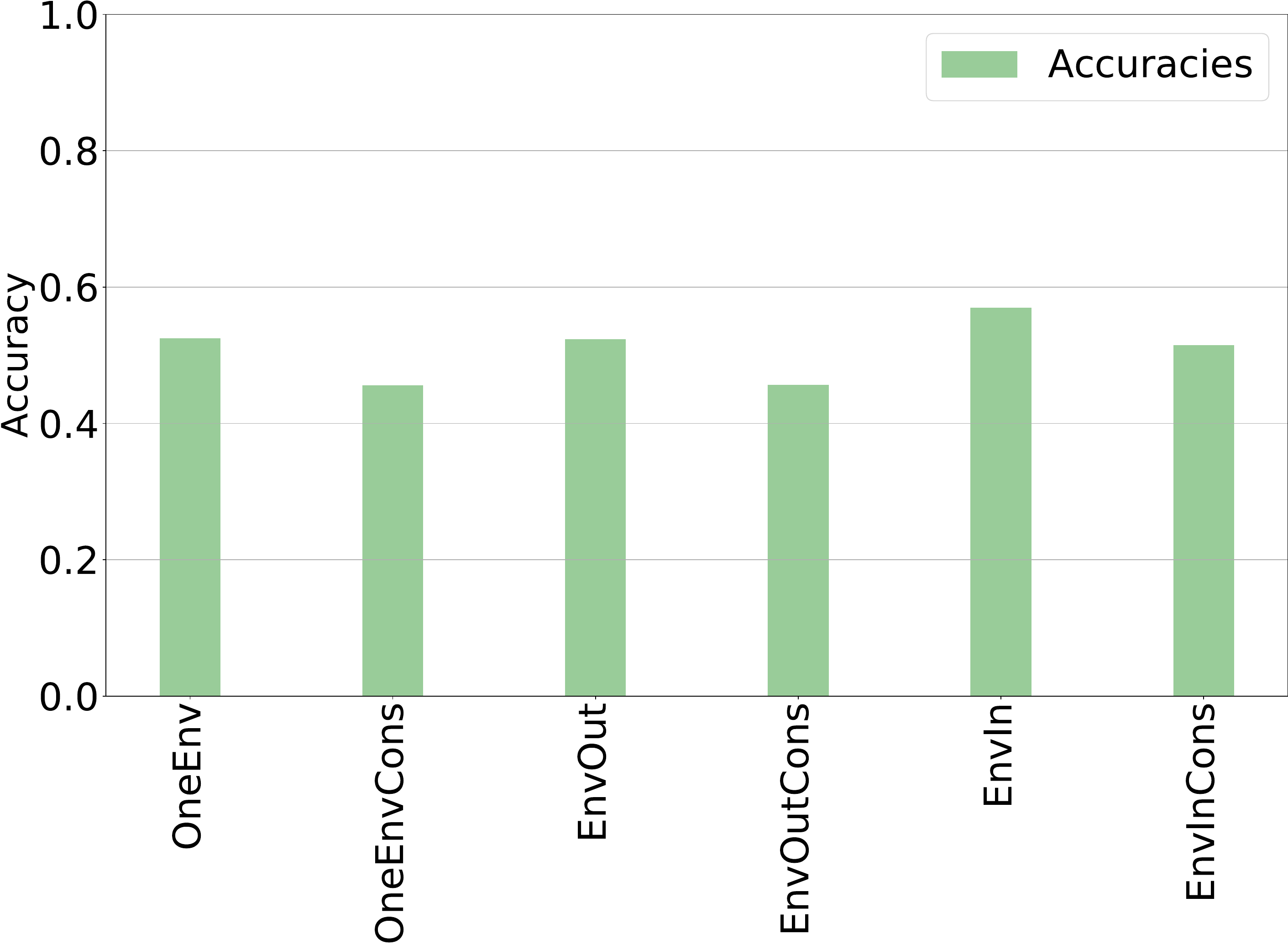}  
    \caption{Detection accuracies of direct causes for different variants of the PC-Algorithm. EnvOut means we pool over all environments and EnvIn means the environment is treated as system intern variable $E$. The suffix Cons means we us the conservative assignment scheme. OneEnv means we only consider the observational environment for inference.}
    \label{fig:pc_alg}
\end{figure}

We furthermore consider three scenarios: (1) the samples across all environments are pooled, (2) only the observational data (from the first environment) is given, and (3) the environment variable is considered as system-intern variable and is seen by the PC-Algorithm (similar as in \citet{mooij2016joint}). Results are shown in Figure \ref{fig:pc_alg}. In order to obtain these results, we sampled $1500$ graphs as described above and applied on each of these datasets a PC-Variant. Best accuracies are achieved if we consider the environment variable as system-intern variable and use the non-conservative orientation scheme (EnvIn).

\subsection{Variable Selection}

We consider the task of finding the direct causes of a target variable $Y$. Our models based on the gating mechanism perform a variable selection and are therefore compared to the PC-Algorithm and ICP. In the following we show the accuracies of this variable selection according to different scenarios. 

\begin{figure}
    \centering
      \includegraphics[width=.8\linewidth]{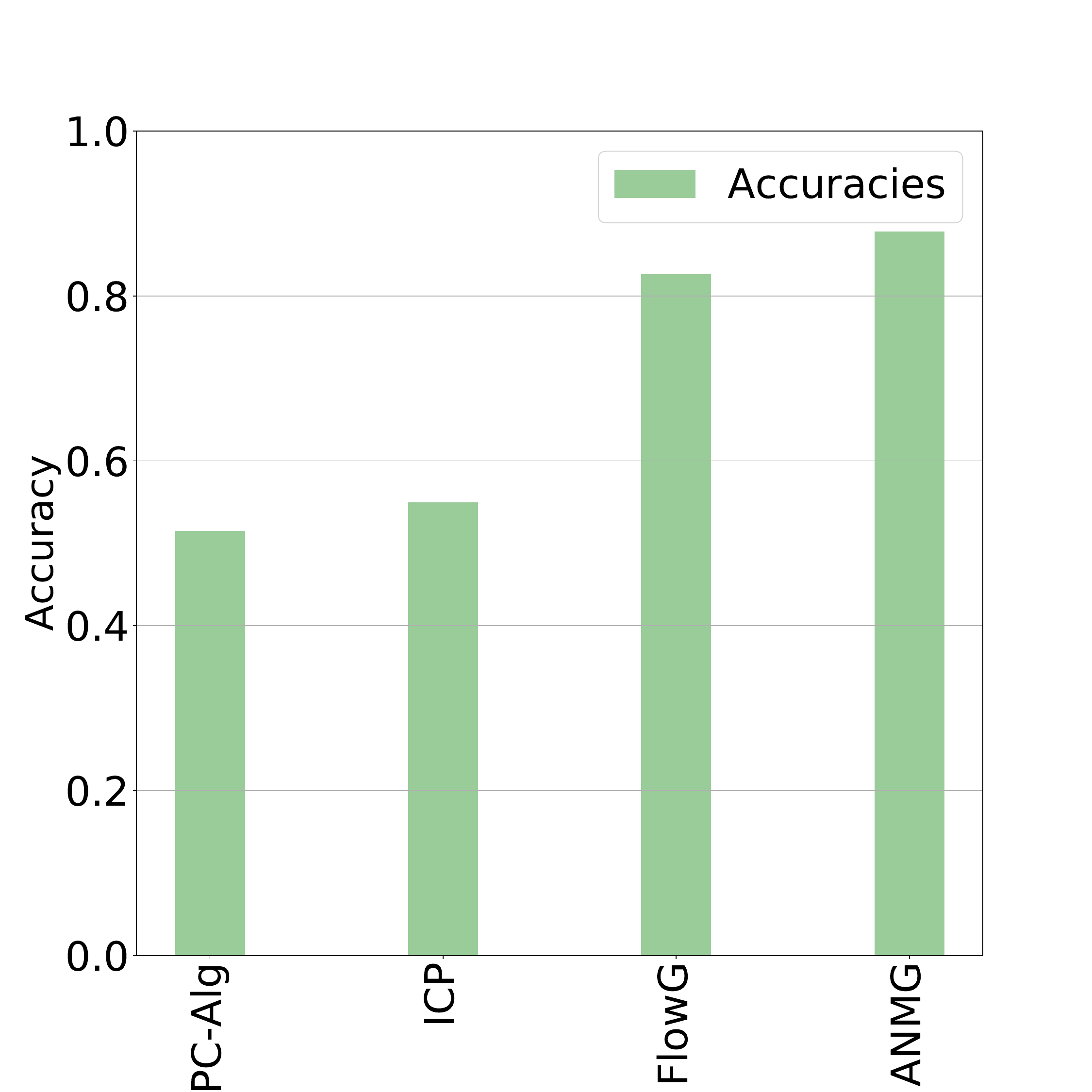}  
    \caption{Accuracies for different models across all scenariso. FlowG and ANMG are our models.}
    \label{fig:all_acc}
\end{figure}

Figure \ref{fig:all_acc} shows the accuracies of ICP, the PC-Algorithm and our models pooled over all scenarios. Our models perform comparably well and better than the baseline in the causal discovery task.

\begin{figure}%[t] %t?
\begin{subfigure}{0.49\textwidth}
  \centering
  % include first image
  \includegraphics[width=.9\linewidth]{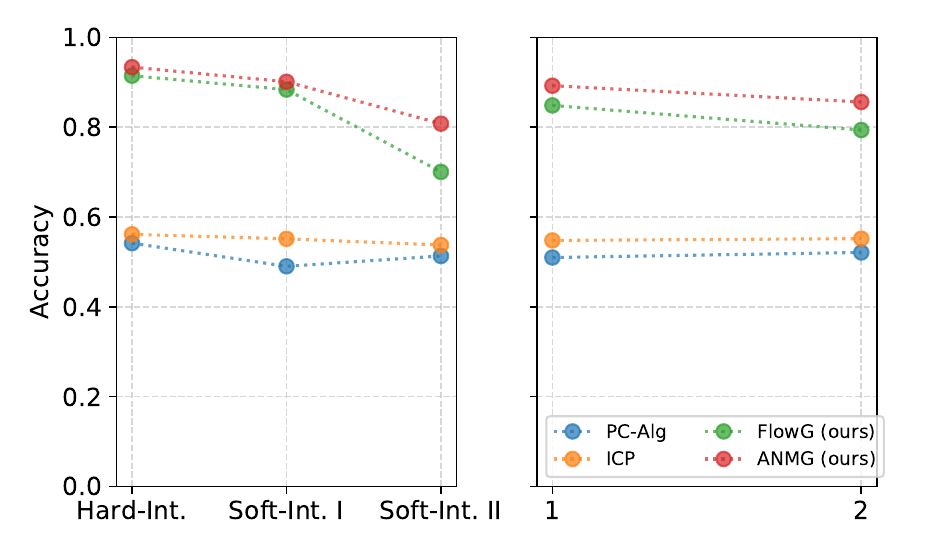}  
  \caption{Accuracies of different models for different intervention types and locations. $1$ stands for intervention on all variables except $Y$ and $2$ stands for interventions on parents and children only.}
  \label{fig:acc_intervention_location_kind}
\end{subfigure}
\begin{subfigure}{0.49\textwidth}
  \centering
  % include second image
  \includegraphics[width=.9\linewidth]{figures/experiment_plots/accuracies_target_mechanism.pdf}  
  \caption{Accuracies of different models according to target variables and mechanisms of the underlying SCM.}
  \label{fig:acc_mechanisms}
\end{subfigure}
\caption{Comparison of models across different scenarios in the causal discovery task.}
\label{fig:heinze_acccuraciesAPP}
\end{figure}

In the following we show results due to different mechanisms, target variables, intervention types and intervention locations.
Figure \ref{fig:acc_mechanisms} shows the accuracies of all models across different target variables. Parentless target variables, i.e. $Y=X_4$ or $Y=X_0$ are easy to solve for ICP due to its conservative nature. All our models solve the parentless case quite well. Performance of the PC-variant depends strongly on the position of the target variable in the SCM indicating that its conservative assignmend scheme has a strong influence on its performance. As expected, the PC-variant deals well with with $Y=X_6$ which is a childless collider. 
The causal discovery task seems to be particularly hard for variable $Y=X_6$ for all other models. This is the variable which has the most parents.

The type of intervention and its location seem to play a minor role as shown in Figure \ref{fig:acc_intervention_location_kind} and Figure \ref{fig:acc_intervention_location_kind}.

Figure \ref{fig:acc_mechanisms} shows that ICP performs well if the underlying causal model is linear, but degrades if the mechanism become non-linear. The PC-Algorithm performs under all mechanisms comparably, but not well. ANMG performs quite well in all cases and even slightly better than FlowG in the cases of additive noise. However in the case of non-additive noise FlowG performs quite well whereas ANMG perform slightly worse -- arguably because their requirements on the underlying mechanisms are not met.

\subsection{Transfer Study}

\begin{figure*}
    \centering
    \includegraphics[width=0.8\linewidth]{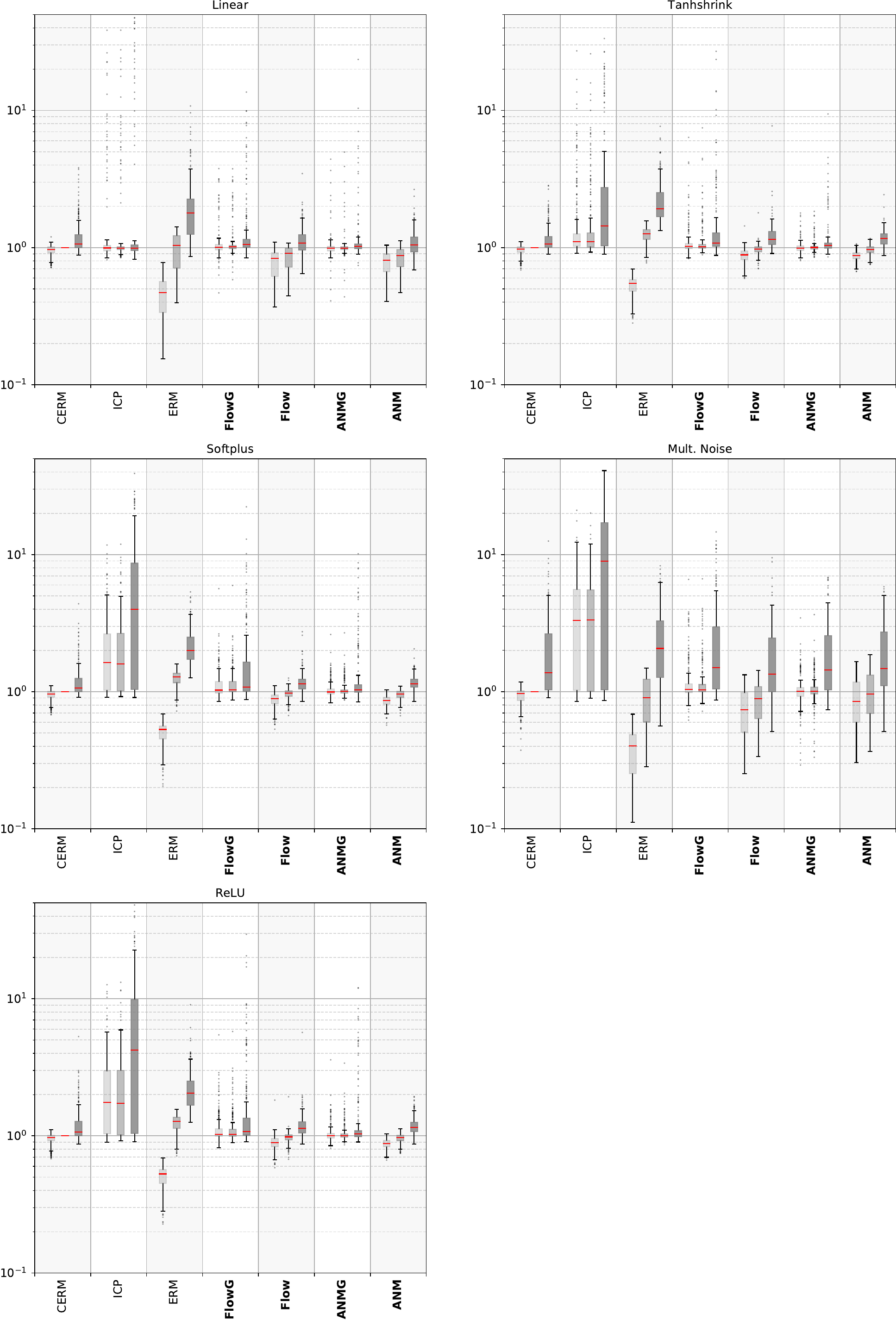}
    \caption{Logarithmic plot of L2 errors, normalized by CERM test error. For each method (ours in bold) from left to right: training error, test error on seen environments, domain generalization error on unseen environments.
    Scenarios for different mechanisms are shown.}
    \label{fig:box_mech}
\end{figure*}

\begin{figure*}
    \centering
    \includegraphics[width=0.8\linewidth]{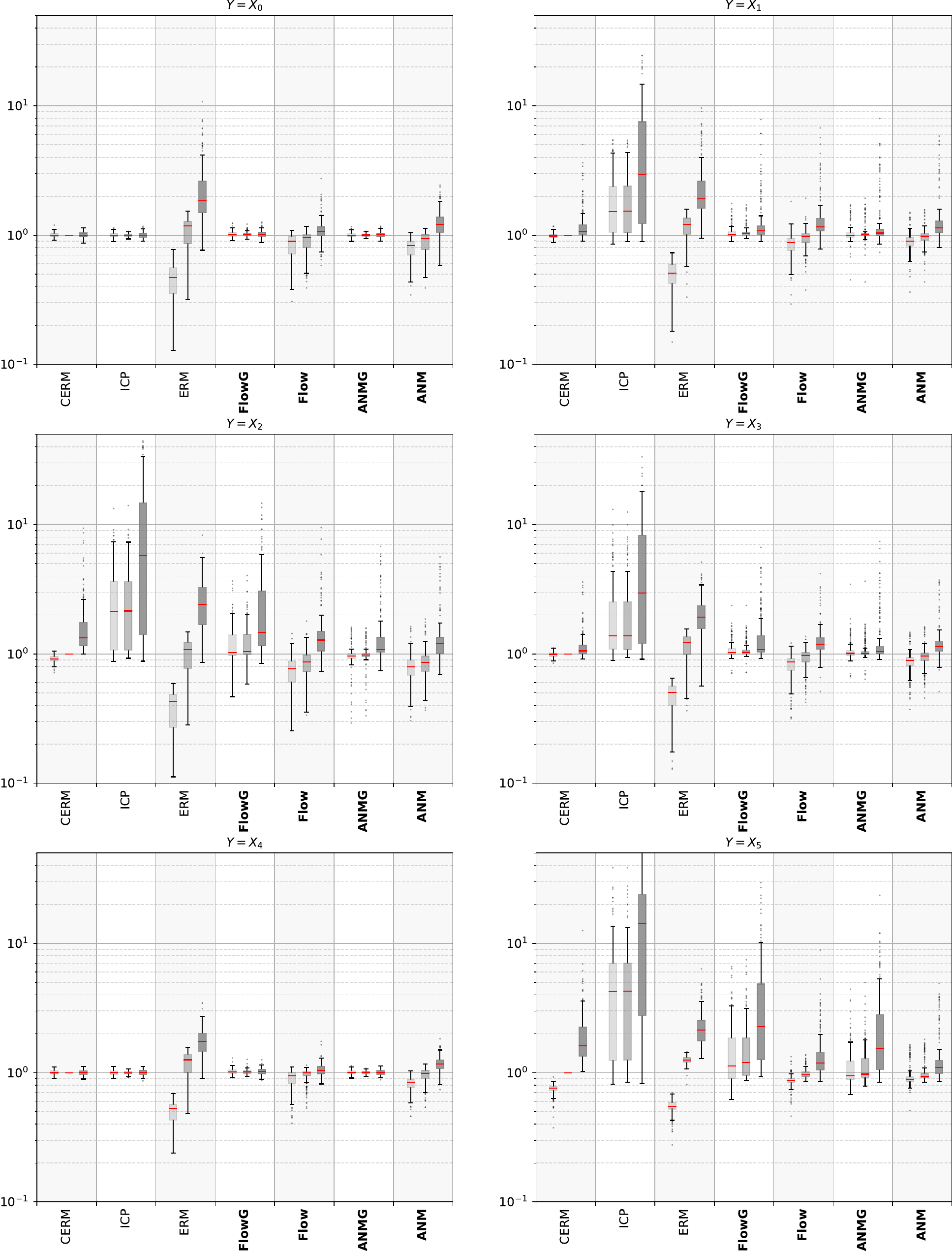}
    \caption{Logarithmic plot of L2 errors, normalized by CERM test error. For each method (ours in bold) from left to right: training error, test error on seen environments, domain generalization error on unseen environments.
    Sceanarios for different target variables are shown.}
    \label{fig:box_target}
\end{figure*}

\begin{figure*}
    \centering
    \includegraphics[width=0.8\linewidth]{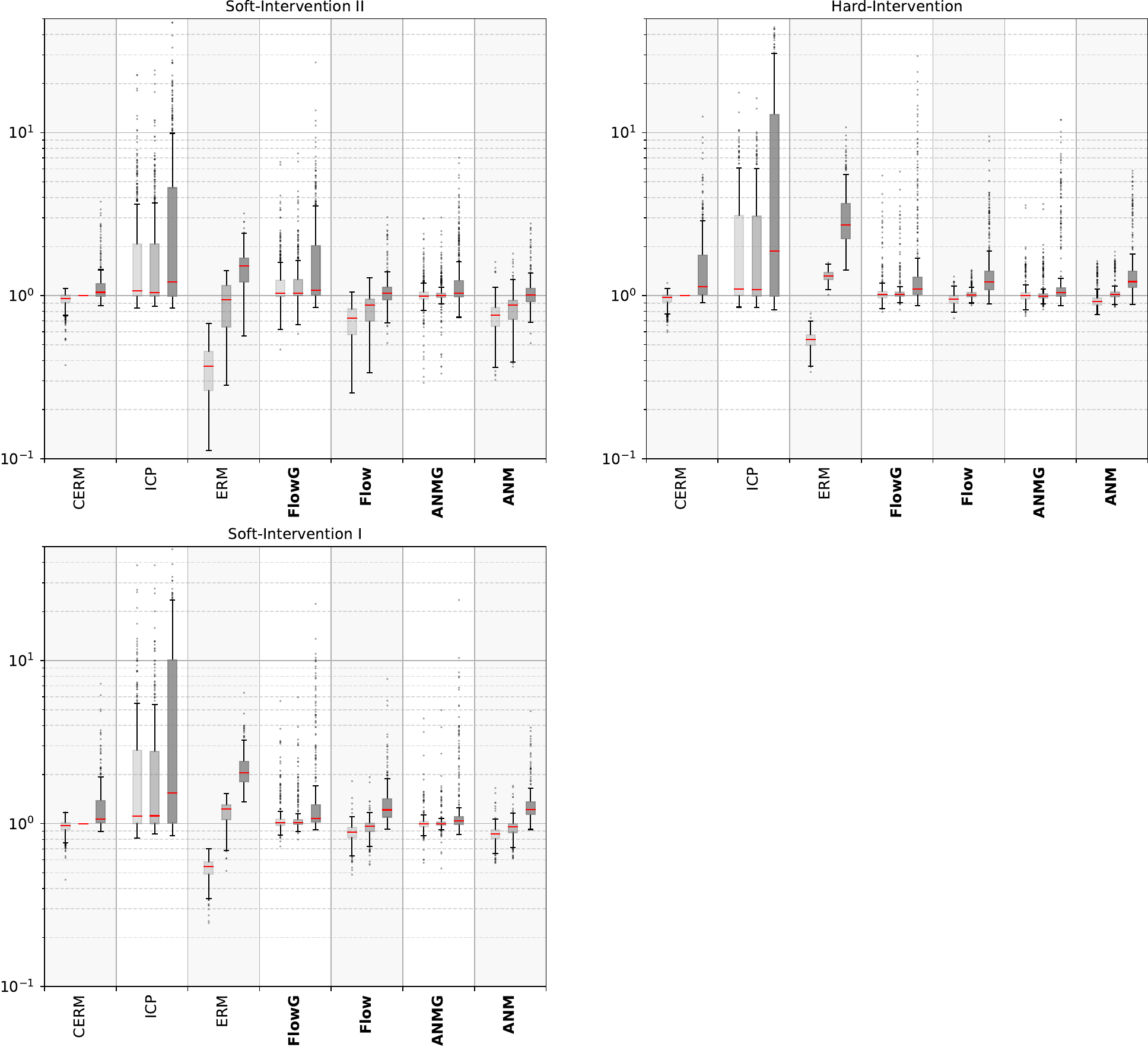}
    \caption{Logarithmic plot of L2 errors, normalized by CERM test error. For each method (ours in bold) from left to right: training error, test error on seen environments, domain generalization error on unseen environments.
    Scenarios for different intervention types are shown.}
    \label{fig:box_intervention_kind}
\end{figure*}

In the following we show the performance of different models on the training set, a test set of the same distribution and a set drawn from an unseen environment for different scenarios. As in Section \ref{sec:experiments}, we use the L2-Loss on samples of an unseen environment to measure out-of-distribution generalization. Figure \ref{fig:box_mech}, \ref{fig:box_target} and \ref{fig:box_intervention_kind} show results according to the underlying mechanisms, target variable or type of intervention respectively. The boxes show the quartiles and the upper whiskers ranges from third quartile to $1.5\cdot IQR$ where $IQR$ is the interquartile range. Similar for the lower whisker.

{}

{}

{}

\section{Experimental Details Colored MNIST}
\label{app:colorMnistExDet}
For the training, we use a feed forward neural network consisting of a feature selector followed by a classificator. The feature selector consists of two convolutional layers with a kernel size of $3$ with $16$ respectively $32$ channels followed by a max pooling layer with kernel size $2$, one dropout layer ($p=0.2$) and a fully connected layer mapping to $16$ feature dimensions. After the first convolutional layer and after the pooling layer a PReLU activation function is applied. %Then we apply a dropout layer with $p=0.2$ and a linear layer mapping to a 16 dimensional vector.
For the classification we use a PReLU activation function followed by a Dropout layer ($p=0.2$) and a linear layer which maps the $16$ features onto the two classes corresponding to the labels.

We use the data generating process from \cite{arjovsky2019invariant}. 50 000 samples are used for training and 10 000 samples as test set. 
For training, we choose a batch size of 1000 and train our models for 60 epochs. We choose a starting learning rate of $6 \cdot 10^{-3}$. The learning rate is decayed by 0.33 after 20 epochs. We use an L2-Regularization loss weighted by $10^{-5}$. After each epoch we randomly reassign the colors and the labels with the corresponding probabilities. The one-dimensional Wasserstein loss is applied dimension-wise and the maximum over dimensions is computed in order to compare residuals. For the HSIC we use a cauchy kernel with $\sigma=1$. The invariance loss $\mathcal{L}_I$ is simply the sum of the HSIC and Wasserstein term.
For Figure \ref{fig:mnist_bar_graph} we trained our model with $\lambda_I \approx 13$. This hyperparameter is chosen from the best run in Figure \ref{fig:mnist_lambda}.

\end{appendix}
\end{document}